\pdfoutput=1
\documentclass[a4paper,fleqn]{cas-dc}
\usepackage[numbers,sort&compress]{natbib}
\usepackage{tabularx}
\usepackage{capt-of}
\usepackage{placeins}
\usepackage{microtype}
\usepackage{algorithm}
\usepackage{algpseudocode}

\renewcommand{\printorcid}{}
\ExplSyntaxOn
\bool_gset_true:N \g_stm_nologo_bool
\ExplSyntaxOff

\ExplSyntaxOn
\cs_set:Npn \__first_footerline:
  {
    \group_begin:
    \small
    \sffamily
    \ifnum\theblind>0\relax
    \else
      \__short_authors: :~
    \fi
    { \rmfamily \itshape Preprint }
    \group_end:
  }
\ExplSyntaxOff

\newcolumntype{Y}{>{\centering\arraybackslash}X}

\newcommand{\metricstdcorner}[2]{#1\kern0.03em\raisebox{-0.45ex}{\normalfont\fontsize{4.5}{4.5}\selectfont\ensuremath{\pm}#2}}

\begin{document}
\let\WriteBookmarks\relax
\def\floatpagepagefraction{1}
\def\textpagefraction{.001}

\shorttitle{}
\shortauthors{Z. Yu et al.}

\title[mode=title]{Dynamic Gaussian Scene Reconstruction by Integrating Events and Structural Priors}

\author[1]{Zihao Yu}
\ead{2120250560@mail.nankai.edu.cn}
\author[1]{Guozhen Liu}
\ead{2120240656@mail.nankai.edu.cn}
\author[1]{Jie Yan}
\ead{2120240546@mail.nankai.edu.cn}
\author[1]{Jialiang Liu}
\ead{2120240614@mail.nankai.edu.cn}
\author[2]{Panlong Tan}
\ead{tanpl@nankai.edu.cn}
\author[1]{Qinglin Sun}
\ead{sunql@nankai.edu.cn}
\author[1]{Hao Sun}\cormark[1]
\ead{sunh@nankai.edu.cn}
\cortext[1]{Corresponding author.}
\affiliation[1]{organization={College of Artificial Intelligence, Nankai University}, city={Tianjin}, postcode={300350}, country={China}}
\affiliation[2]{organization={Haihe Laboratory of Information Technology Application Innovation}, city={Tianjin}, postcode={300450}, country={China}}

\begin{abstract}
Existing approaches to event-based dynamic Gaussian reconstruction rely on two assumptions for support augmentation and reliability estimation. First, existing Gaussians can absorb reconstruction errors. Second, photometric optimization has converged sufficiently to estimate reliability using analytical variance. Both assumptions can fail under fast motion. When the first fails, regions without Gaussian support cannot generate densification candidates. When the second fails, Gaussians in moving regions where optimization has not converged may be excluded as motion references. We propose Event--Structure Collaborative Dynamic Gaussian Reconstruction (ESC-DGR), using event observations independent of both conditions. Events first identify changes that require additional support. This support is placed only where rendered depth is valid. The augmented representation still needs reliable motion references. Online rendering statistics and historical event evidence therefore propagate displacement estimates from reliable Gaussians to weakly observed neighborhoods. These references describe dominant motion at the frame-interval timescale. Target-time events therefore further condition bounded position residuals for active Gaussians. Support placement, reference selection, and residual prediction thus operate on the same evolving representation within one optimization run. For event-driven monocular dynamic scene reconstruction, our method achieves the best reconstruction quality among the evaluated methods on the E-D3DGS benchmark. Relative to E-D3DGS, dynamic-region PSNR (dPSNR) improves by 1.02 and 1.45~dB on synthetic and real scenes, respectively. The total dynamic Gaussian count across eight synthetic scenes decreases by 8.5\%. Although mean frame rate decreases by about 21\%--25\% relative to E-D3DGS, rendering remains above 80~FPS in all twelve scenes. The implementation is publicly available at \url{https://github.com/zh-yu1/ESC-DGR}.
\end{abstract}


\begin{keywords}
Dynamic scene reconstruction \sep Event cameras \sep Uncertainty-aware optimization \sep Graph propagation \sep Multimodal fusion \sep Residual networks
\end{keywords}

\maketitle

\hypersetup{
  pdftitle={Dynamic Gaussian Scene Reconstruction by Integrating Events and Structural Priors},
  pdfauthor={Zihao Yu, Guozhen Liu, Jie Yan, Jialiang Liu, Panlong Tan, Qinglin Sun, Hao Sun},
  pdfsubject={Event-based monocular dynamic scene reconstruction with 3D Gaussian splatting},
  pdfkeywords={Dynamic scene reconstruction, Event cameras, Uncertainty-aware optimization, Graph propagation, Multimodal fusion, Residual networks},
  pdfcreator={LaTeX with cas-dc}
}

\section{Introduction}

Dynamic scene reconstruction from monocular observations supports immersive interaction, content creation, and robotic perception. It has also recently been combined with online point tracking~\cite{ref49}. We adopt monocular capture to simplify acquisition and reduce deployment costs. This avoids the additional inter-camera calibration and synchronization required by stereo systems. Stereo provides disparity-based depth constraints, but its correspondences remain vulnerable to fast motion, occlusion, and weak texture. Additional viewpoints also cannot recover temporal changes absent from RGB frames. Without simultaneous multi-view constraints, however, monocular geometry and motion estimation are much more underconstrained. Under sparse observations, this produces drift, ghosting, and discontinuous motion boundaries. We consider short monocular sequences with known camera poses and temporally aligned RGB and event observations.

Event cameras record changes in log intensity asynchronously, providing temporal cues between RGB frames. Events have been used to recover Gaussian representations and camera trajectories from blurred images~\cite{ref32}. Their effective use nevertheless requires sufficient geometric support and reliable motion constraints. Fast-moving objects, thin structures, and briefly visible regions often have sparse initial support, while independent deformation of weakly observed Gaussians can cause local drift.

Existing support augmentation and reliability estimation rely on two implicit assumptions. The first is that reconstruction errors can be absorbed by existing Gaussians. Most event-based dynamic methods retain 3DGS densification: accumulated screen-space positional gradients trigger cloning and splitting of existing primitives~\cite{ref1}, while events mainly provide supervision, initialization, or threshold calibration (Section \ref{sec:related-event}). TAD-GS adjusts gradients and thresholds~\cite{ref44}, and MoDGS constrains geometry using depth priors~\cite{ref50}, but neither changes this source of candidates. VAD-GS generates support in missing regions through complementary views and stereo matching~\cite{ref43}, which are not directly available in our monocular setting. When fast motion extends beyond initial coverage, errors arise where no Gaussian exists to provide a positional gradient. Nearby Gaussians must instead absorb these errors, potentially undergoing repeated splitting while the missing regions remain unsupported. The difficulty is therefore not only how strongly to densify, but where new support can originate.

The second assumption is that photometric optimization has converged. Occlusion, weak texture, and fast motion leave some Gaussians observed in only a few frames, so their motion cannot be treated as equally reliable. Existing methods assess reliability from external correspondences~\cite{ref48,ref47}, Bayesian uncertainty~\cite{ref39}, or rendering statistics~\cite{ref24}. In particular, USplat4D's analytical variance approximates underlying Gaussian uncertainty when the loss is near a local minimum and predictions over the covered pixels have converged. Its two-stage procedure further optimizes a pretrained dynamic model to satisfy this condition. In our single-stage setting, however, the Gaussian set and supervision continue to change as training proceeds. Fast-moving regions may remain unconverged, so applying the same criterion can classify them collectively as unreliable and remove precisely the motion references needed in those regions.

These assumptions share a difficulty: the evidence used to assess the representation is derived from the representation itself. Densification relies on existing Gaussians to identify missing support. Convergence-dependent reliability estimation assesses their motion using rendering results that are still being optimized. Events arise from observed intensity changes regardless of whether the corresponding region contains Gaussians or whether rendering has converged. They can therefore locate changes without requiring positional gradients and provide external evidence alongside accumulated rendering statistics. This motivates treating support placement and motion-reference selection as related decisions within the same evolving representation.

Resolving these decisions does not remove the temporal limitation. Even with sufficient support and reliable references, frame-level deformation and interpolation between adjacent reference states may not explain rapid motion between those states. Averaging such displacements produces blurred boundaries and ghosting. Events provide the only sub-frame observations in our setting, but their sparsity and noise make them insufficient to recover the full motion independently. We therefore use target-time events to refine the motion already supported by the Gaussian geometry and structural references. Building on E-D3DGS, ESC-DGR follows this reasoning within one optimization run, using a shared dynamic Gaussian set rather than introducing another scene representation.

\begin{figure}[width=\columnwidth,pos=htbp]
  \centering
  \includegraphics[width=\columnwidth,trim=0 4bp 0 0,clip]{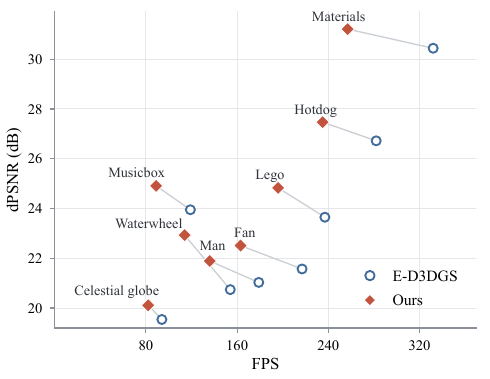}
  
  \caption{Frame rate and dPSNR across eight synthetic scenes. Blue open circles denote E-D3DGS, and orange-red diamonds denote our method. Thin lines connect the two results for each scene, with text labels indicating scene names. Data are from Tables \ref{tab:2} and \ref{tab:dynamic-region}. Section \ref{sec:experiment-settings} describes the test protocol with synchronized events.}
  \label{fig:speed-quality}
\end{figure}

Fig. \ref{fig:speed-quality} shows the quality--speed trade-off on eight synthetic scenes. Mean dPSNR increases from 23.46 to 24.48~dB, while mean rendering speed decreases from about 202 to 159~FPS (approximately 21\%). dPSNR improves in every scene, and rendering speed remains above 80~FPS in all scenes.

Our main contributions are summarized as follows:

\begin{itemize}
  \item We identify two assumptions that can fail under fast motion: existing Gaussians can absorb reconstruction errors, and photometric optimization has converged. To address the resulting support gaps, Event-guided Dynamic Support Augmentation (EDSA) locates additional support using temporally fused events and valid rendered depth, without requiring gradients from Gaussians in uncovered regions.

  \item To provide motion references for the augmented representation without assuming photometric convergence, we introduce the Contribution-Confidence Structural Motion Graph (CCSMG). Online rendering statistics and historical event evidence select reliable reference Gaussians whose displacements guide weakly observed neighborhoods, without external trackers or additional supervision.

  \item We introduce Event-conditioned High-frequency Motion Refinement (EHMR) to address rapid inter-frame changes that remain beyond the structural motion estimate. Current events determine where refinement is needed, while contribution reliability and graph context condition bounded position residuals for active Gaussians.

  \item For event-driven monocular dynamic scene reconstruction, ESC-DGR achieves state-of-the-art reconstruction quality on the twelve-scene E-D3DGS benchmark. Synthetic and real dPSNR improve by 1.02 and 1.45~dB, respectively; the synthetic dynamic Gaussian count falls by 8.5\%, with rendering above 80~FPS in every scene.
\end{itemize}

\section{Related work}

\subsection{Dynamic scene representations and motion modeling}

Neural radiance fields connect scene representations to image observations through continuous coordinate queries and volume rendering~\cite{ref18}. Early dynamic methods represent motion using canonical-space deformation or explicit spatiotemporal encoding~\cite{ref3,ref15,ref4}. 3DGS achieves real-time rendering with anisotropic Gaussians and differentiable rasterization~\cite{ref1}, while 4DGS and Deformable-3DGS extend this representation to dynamic reconstruction through spatiotemporal features and deformation fields~\cite{ref6,ref7}. SC-GS and Gaussian-Flow organize per-Gaussian motion using sparse control and explicit temporal functions, respectively~\cite{ref37,ref38}.

Since 2025, research has increasingly focused on motion hierarchies, continuous time, and local modeling of complex motion. MoDec-GS compresses dynamic representations through global-to-local motion decomposition and temporal interval adjustment~\cite{ref57}. HiMoR organizes coarse-to-fine deformation using a tree hierarchy and shared motion bases~\cite{ref58}. SplineGS describes continuous trajectories with motion-adaptive splines and jointly optimizes camera parameters~\cite{ref59}. In 2026, RetimeGS addresses temporal aliasing in frame interpolation through explicit temporal behavior, optical flow, and rendering supervision at multiple times~\cite{ref60}. MAPo groups Gaussians by motion intensity, subdivides temporal intervals for highly dynamic parts, and assigns dedicated deformation networks~\cite{ref62}. Although hierarchical structures and continuous-time constraints improve complex motion modeling, sparse observations still leave two questions: whether geometric support is sufficient and which Gaussians can provide reliable local motion references.

\subsection{Support and observation reliability in dynamic Gaussians}

Dynamic Gaussian support is established through density control and geometric initialization. In 3DGS, accumulated screen-space positional gradients trigger cloning and splitting of existing Gaussians. Regions without Gaussians therefore cannot generate candidates~\cite{ref1}. TAD-GS weights densification gradients by temporal visibility and adjusts thresholds~\cite{ref44}. VAD-GS selects complementary views using voxel visibility and generates new Gaussians in missing regions through stereo matching~\cite{ref43}. MoDGS constrains geometry with monocular depth priors~\cite{ref50}. These methods rely on gradients, visibility, and depth priors, respectively, but do not use the dense change observations provided by events to locate new Gaussians during optimization. Insufficient coverage of fast motion and thin structures therefore remains difficult to address under sparse monocular observations.

Motion constraints and observation reliability also affect reconstruction stability. Deformation-field methods typically apply uniform regularization to all Gaussians. Examples include the elastic regularization of Nerfies and the annealed temporal regularization of Deformable-3DGS~\cite{ref3,ref7}. Such uniform regularization makes it difficult to distinguish reliable regions from weakly constrained ones. Shape of Motion uses tracker-derived visibility and uncertainty to reduce the influence of unreliable correspondences~\cite{ref48}. MoSca propagates motion to unobserved locations using trajectory visibility, epipolar errors, and local rigidity~\cite{ref47}. Bayes' Rays and USplat4D estimate per-point and per-Gaussian uncertainty using Bayesian inference and the variance of alpha-compositing weights, respectively~\cite{ref39,ref24}. In 2026, GP-4DGS further models spatiotemporal deformation correlations with variational Gaussian processes to estimate motion uncertainty and predict motion in weakly observed regions~\cite{ref61}. These approaches rely on external correspondences, probabilistic motion models, or rendering statistics. We focus on online motion-reference selection during single-stage joint event--RGB training. In this setting, ongoing optimization makes the photometric convergence required by USplat4D's analytical variance difficult to maintain.

\subsection{Event-driven reconstruction and dynamic Gaussians}
\label{sec:related-event}

Event cameras independently monitor per-pixel log intensity and asynchronously emit signed events when positive or negative thresholds are reached~\cite{ref9}. Their sparse output depends on scene activity, pixel count, and threshold settings, with microsecond temporal resolution and dynamic range exceeding 120~dB. Events encode changes rather than absolute intensity, and thresholds vary across pixels and operating conditions.

Event representations are moving from generic temporal encoding toward joint design with imaging tasks. A task-oriented 3D Gaussian event representation proposed in 2026 adaptively selects spatiotemporal coordinates according to image content and event density, then aggregates local motion features using Gaussian weights for motion deblurring~\cite{ref65}. These Gaussians are spatiotemporal aggregation kernels for events, distinct from the geometric Gaussians of a 3D scene. Recovering dynamic geometry from event features still requires projection relationships and motion constraints across time.

EventNeRF and E-NeRF established event-based radiance field reconstruction, while E2NeRF extended this approach by combining blurred RGB images with events~\cite{ref11,ref10,ref13}. Recent work increasingly optimizes Gaussian representations directly. EvaGaussians incorporates event streams into Gaussian initialization and optimization from blurred images~\cite{ref32}. DiET-GS combines event double-integral constraints with diffusion priors to improve color and edge details~\cite{ref63}. EventSplat uses an event-to-video prior for initialization and camera trajectory splines to handle fast camera motion~\cite{ref51}. Introduced in 2026, E2EGS extracts edges from local temporal event consistency to guide Gaussian initialization and joint optimization with unknown camera poses~\cite{ref64}. These studies show that events can guide geometric initialization as well as loss supervision. Our focus is on augmenting the dynamic set with trainable support after static--dynamic decomposition and constraining its motion throughout optimization. E-NeMF further uses events to constrain a parameterized motion field, enabling novel spatiotemporal view synthesis of dynamic scenes~\cite{ref52}.

Events have only recently been combined with dynamic Gaussians. DEGS establishes geometric correspondences through event optical flow and supervises deformation fields with pseudo-labels~\cite{ref41}. FastEventDGS reconstructs fast dynamic scenes from a single event camera using continuous camera trajectories and a local event motion loss~\cite{ref42}. E-4DGS optimizes dynamic Gaussians from multi-view event streams through event-based initialization and learnable contrast thresholds~\cite{ref45}. STD-GS obtains a spatiotemporally decoupled representation through frame--event clustering, with event features also used for clustering during inference~\cite{ref46}. E-D3DGS is most closely related to our work. It jointly models Gaussians and thresholds, uses RGB intensity changes to assist event threshold estimation, and advances event-based dynamic reconstruction through soft static--dynamic decomposition and a public benchmark~\cite{ref14}. These methods emphasize event motion priors, initialization, threshold modeling, or frame--event interaction. Building on them, we study the interaction among support augmentation, reliable motion references, and local refinement. After static--dynamic decomposition, we add trainable dynamic support, construct motion references from online observation statistics, and predict bounded position residuals conditioned on events at the target time.

The methods discussed above do not jointly address insufficient dynamic support, motion ambiguity in weakly observed Gaussians, and rapid local inter-frame changes within a single framework. We use dense temporal event cues to address all three on a shared dynamic Gaussian set and evaluate the resulting framework on a public event-based dynamic reconstruction benchmark.

\section{Preliminaries}

\subsection{Dynamic Gaussian parameterization and differentiable rendering}

A canonical scene consists of $N$ anisotropic Gaussians~\cite{ref1}, where $N$ is the number of Gaussians in the current representation. The $i$-th Gaussian is denoted by $G_i$:

\begin{equation}
\mathcal G=\{G_i\}_{i=1}^{N},\quad G_i=(\boldsymbol\mu_i,\mathbf R_i,\mathbf S_i,\alpha_i,\mathbf c_i)
\label{eq:1}
\end{equation}

Here, $\boldsymbol\mu_i$, $\mathbf R_i$, $\mathbf S_i$, $\alpha_i$, and $\mathbf c_i$ denote center, rotation, positive axial scales, opacity, and view-dependent spherical-harmonic color, respectively. The covariance is $\boldsymbol\Sigma_i=\mathbf R_i\mathbf S_i\mathbf S_i^{\mathsf T}\mathbf R_i^{\mathsf T}$. Appendix \ref{app:gaussian} gives the spatial kernel and projection conventions. A deformation field maps canonical Gaussians to time $t$~\cite{ref6,ref7}. Our low-frequency branch predicts translation, rotation, and scale offsets; the high-frequency branch adds position residuals. Depth-sorted alpha compositing gives

\begin{equation}
\begin{aligned}
\widehat I(\mathbf p,t)&=\sum_{i\in\mathcal N(\mathbf p,t)}T_i(\mathbf p,t)\widetilde\alpha_i(\mathbf p,t)\mathbf c_i(t),\\
T_i(\mathbf p,t)&=\prod_{j<i}[1-\widetilde\alpha_j(\mathbf p,t)].
\end{aligned}
\label{eq:3}
\end{equation}

Here, $\mathcal N(\mathbf p,t)$ contains Gaussians covering pixel $\mathbf p$, sorted by depth. The effective alpha $\widetilde\alpha_i=\alpha_i\varphi_i^{\mathrm{2D}}$ combines opacity with the projected kernel, and $T_i$ is the transmittance before Gaussian $i$. Thus, $j<i$ indexes preceding Gaussians along the ray; in the motion graph below, $j$ indexes adjacent reference Gaussians.

\subsection{Event representation and polarity integration}

We bin events over time and combine them into event activity and feature maps (Sections \ref{sec:prior} and \ref{sec:residual}), allowing asynchronous observations to guide Gaussian optimization. To use events for Gaussian reconstruction, we represent a single event as

\begin{equation}
e_n=(x_n,y_n,t_n,p_n),\quad p_n\in\{-1,+1\}
\label{eq:4}
\end{equation}

Here, $n$ indexes events, and $(x_n,y_n)$, $t_n$, and $p_n$ denote pixel position, timestamp, and polarity, respectively. At a fixed pixel $\mathbf p$, the discrete events define an impulse stream in continuous time. With $c_{\phi}$ denoting the contrast threshold controlled by parameters $\phi$, the intensity change observed through events over $[t_a,t_b]$ is
\begin{equation}
\begin{aligned}
e_{\mathbf p}(\tau)
&=\sum_{n:(x_n,y_n)=\mathbf p}
p_n\delta_{\mathrm D}(\tau-t_n),\\
E_{a,b}(\mathbf p)
&=\int_{t_a}^{t_b}
c_{\phi}(\mathbf p,\tau)
e_{\mathbf p}(\tau)\,\mathrm d\tau.
\end{aligned}
\label{eq:event-integral}
\end{equation}
Here, $\delta_{\mathrm D}$ is the Dirac delta, distinct from the position residual $\boldsymbol\delta$ below. The integral is implemented as a sum of signed thresholds. E-D3DGS calibrates $c_\phi$ from training RGB intensity changes and compares event integrals with rendered log-intensity changes (Section \ref{sec:objective}).

Training uses sparse RGB frames $\{(I_f,\Pi_f,t_f)\}_{f=1}^{F}$ and event packets between adjacent frames. Here, $F$ is the number of RGB frames and $f$ is the frame index. The variables $I_f$, $\Pi_f$, and $t_f$ denote the corresponding image, camera intrinsics and extrinsics, and sampling time. The goal is to learn a static Gaussian set $\mathcal G^s$, a dynamic Gaussian set $\mathcal G^d$, and temporal deformation in canonical space such that renderings at both training and interpolated times are consistent with RGB and event observations. Superscripts $s$ and $d$ denote the static and dynamic components, respectively.

\section{Method}
\subsection{Framework overview}
\begin{figure*}[width=\textwidth,pos=t]
\centering
\noindent\makebox[\textwidth][c]{%
  \includegraphics[width=0.98\textwidth,trim=0 13bp 0 0,clip]{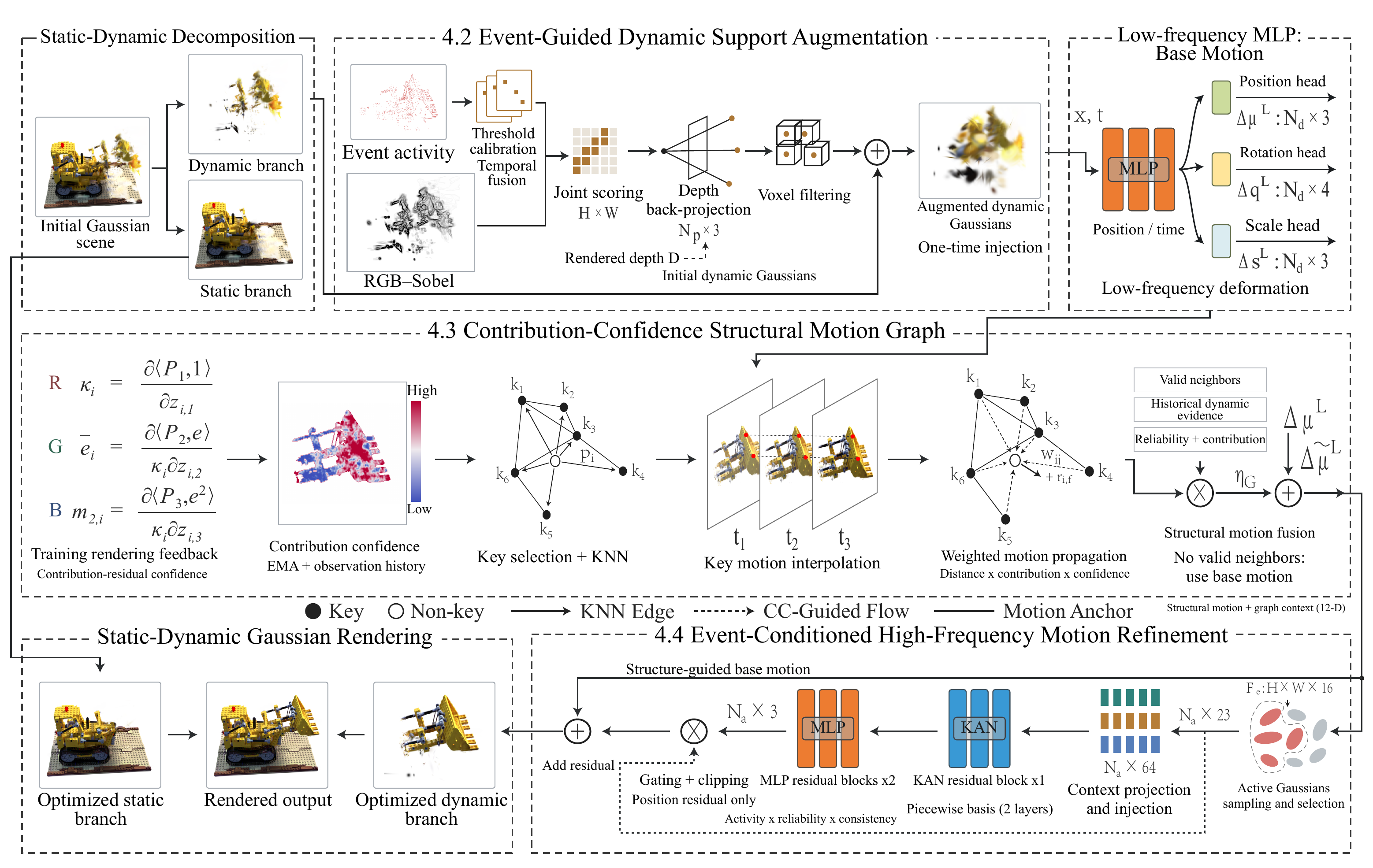}}
\caption{Overview of ESC-DGR. Event activity locates support beyond gradient-generated candidates; online contribution confidence provides motion references without assuming photometric convergence. Target-time events refine the remaining local displacements using the resulting geometry and graph context. The static and deformed dynamic branches are rendered jointly. Dynamic support is added once, while contribution confidence is updated periodically from rendering feedback during training. Thumbnails, graph structures, and displacement arrows provide schematic illustrations of the mechanisms.}
\label{fig:1}
\end{figure*}
Fast motion breaks the connection between reconstruction error and the evidence used to correct it. A region without Gaussians cannot generate candidates through their positional gradients. A region where optimization has not converged cannot supply the converged photometric predictions required to estimate reliability using analytical variance. As shown in Fig. \ref{fig:1}, ESC-DGR addresses these failures within the evolving E-D3DGS representation. We first use event activity to locate missing support (Section \ref{sec:prior}). Once this support is available, its motion remains unevenly constrained. Motion references must therefore be selected using online observation evidence rather than a converged model (Section \ref{sec:structure}). Target-time events then refine displacements that the resulting frame-level motion cannot explain (Section \ref{sec:residual}).

Static--dynamic decomposition gives a static Gaussian set $\mathcal G^s$ and an initial dynamic set $\mathcal G_0^d$. Events indicate candidate locations, while valid rendered depth places them in the current scene geometry. The supplementary set $\mathcal G_{\mathrm{sup}}^d$ is added once:
\begin{equation}
\mathcal G^d\leftarrow\mathcal G_0^d\cup\mathcal G_{\mathrm{sup}}^d.
\label{eq:support-union}
\end{equation}
New and existing Gaussians share the same motion model and reconstruction objective. Adding support supplies primitives that can explain the observations, but does not determine their motion. We therefore estimate contribution confidence during optimization and use reliable reference Gaussians to guide the low-frequency displacements of weakly observed neighbors. The resulting motion estimate also determines where event features are sampled for local refinement. Support, confidence, and event conditioning are thus connected through the states of the same Gaussians.

Let $\boldsymbol\mu_i^0$ denote the canonical-space center, $\Delta\boldsymbol\mu_{i,f}^L$ the base low-frequency displacement at frame $f$, and $\Delta\boldsymbol\mu_{i,f}^G$ the reference displacement obtained through graph propagation. Structural guidance is written as
\begin{equation}
\begin{aligned}
\widetilde{\Delta\boldsymbol\mu}_{i,f}^{L}
={}&\Delta\boldsymbol\mu_{i,f}^{L}
+\eta_G(k)b_{i,f}^{G}\\
&\cdot\left(\Delta\boldsymbol\mu_{i,f}^{G}
-\operatorname{sg}[\Delta\boldsymbol\mu_{i,f}^{L}]\right),
\end{aligned}
\label{eq:decomp}
\end{equation}
Here, $\operatorname{sg}$ denotes stop-gradient, and $\eta_G(k)$ is a guidance coefficient that gradually increases with training iteration $k$. The gate $b_{i,f}^{G}\in\{0,1\}$ determines whether a Gaussian receives structural guidance based on neighborhood validity, reliability, and historical dynamic evidence. Details are given in Section \ref{sec:structure}. The final dynamic center is
\begin{equation}
\boldsymbol\mu_{i,f}=\boldsymbol\mu_i^0+
\widetilde{\Delta\boldsymbol\mu}_{i,f}^{L}+\boldsymbol\delta_{i,f}^{\mathrm{hi}},
\label{eq:final-position}
\end{equation}
Here, $\boldsymbol\delta_{i,f}^{\mathrm{hi}}$ is the event-conditioned bounded position residual. Equation \eqref{eq:decomp} preserves the base low-frequency displacement when no valid graph reference is available, and $\boldsymbol\delta_{i,f}^{\mathrm{hi}}$ is zero for inactive Gaussians. Rotation and scale remain modeled by the low-frequency branch. Deformed dynamic Gaussians and static Gaussians jointly receive RGB and event supervision. Throughout this paper, $t$ denotes scene time, $k$ denotes the training iteration, and $u$ indexes valid-observation updates. Bold symbols denote vectors or matrices. Calligraphic set symbols and scalars are defined separately.

\subsection{Event-guided Dynamic Support Augmentation (EDSA)}
\label{sec:prior}

The first assumption fails when the error occurs outside the existing Gaussian coverage. For a rapidly moving thin object, a few Gaussians may cover its boundary while leaving much of its surface unrepresented. Event supervision still reaches the model through the rendering contributions of existing Gaussians. Consequently, large errors in the missing regions may increase nearby positional gradients and trigger further splitting without creating candidates where support is absent. Repeated subdivision can fragment the represented geometry while leaving the gaps unresolved.

We use events to determine candidate locations independently of these gradients. EDSA fuses event activity over time, scores changing pixels, and back-projects them using valid rendered depth after static--dynamic decomposition. This relaxes the dependence of candidate generation on existing dynamic Gaussians, although it does not remove the need for valid scene depth. We hypothesize that placing support in these regions can distribute reconstruction errors more broadly and reduce subsequent densification pressure. Gaussian counts and reconstruction quality provide observable tests of this hypothesis; the redistribution of errors itself is not directly measured.

\subsubsection{Temporal fusion and candidate scoring}

Temporal fusion suppresses isolated events but can also weaken newly emerging motion. We therefore combine fused activity with its difference from current activity, treating persistent responses and pronounced instantaneous changes as more reliable than intermediate differences. Appearance boundaries further guide candidate ranking. These cues locate support rather than serve as supervision targets.

Let $N_{f,b}^{+}$ and $N_{f,b}^{-}$ denote the positive and negative event count maps in temporal bin $b$ of event packet $f$. The calibrated thresholds $\widehat c_{f,b}^{\mathrm{thr},\pm}$ are supplied by a pretrained threshold network. We first take the absolute value of the calibrated signed event accumulation and then fuse it within a local temporal window $\mathcal T_f$:
\begin{equation}
\begin{aligned}
A_f(\mathbf p)&=\left|\sum_b\widehat c_{f,b}^{\mathrm{thr},+}N_{f,b}^{+}(\mathbf p)
-\sum_b\widehat c_{f,b}^{\mathrm{thr},-}N_{f,b}^{-}(\mathbf p)\right|,\\
\bar A_f&=\frac{\sum_{j\in\mathcal T_f}\omega_{fj}A_j}
{\sum_{j\in\mathcal T_f}\omega_{fj}},\quad
\omega_{fj}=r+1-|j-f|.
\end{aligned}
\label{eq:6}
\end{equation}
The set $\mathcal T_f=\{j\in\{1,\ldots,F\}:|j-f|\le r\}$ contains neighboring frame indices, where $r$ is the temporal window radius. Triangular weights give nearby observations greater influence on current activity. Using the maximum fused activity within the frame as a reference, we define the event-support candidate mask as
\begin{equation}
M_f^{\mathrm{sup}}(\mathbf p)=
\mathbb I[\bar A_f(\mathbf p)>\tau_{\mathrm{sup}}\max_{\mathbf p'}\bar A_f(\mathbf p')],
\label{eq:7}
\end{equation}
Here, $\tau_{\mathrm{sup}}$ is a relative threshold, and $\mathbb I$ is the indicator function. This mask determines candidate pixel locations for Gaussian support.

Let $\widetilde D_f^{\mathrm{temp}}$ be the within-frame min--max normalization of $|\bar A_f-A_f|$, set to 1 for a constant map. The activity $\widetilde A_f$ uses the same normalization. Let $g_f\in[0,1]$ denote normalized Sobel magnitude from grayscale RGB. Appendix \ref{app:normalization} gives these operations. To favor both persistent responses and pronounced instantaneous changes over ambiguous intermediate differences, we define
\begin{equation}
\begin{aligned}
r_f^{\mathrm{temp}}(\mathbf p)
&=\max\{\widetilde D_f^{\mathrm{temp}}(\mathbf p),1-\widetilde D_f^{\mathrm{temp}}(\mathbf p)\},\\
\psi_f(\mathbf p)
&=\widetilde A_f(\mathbf p)
\bigl[(1-\lambda_g)+\lambda_g g_f(\mathbf p)\bigr]\\
&\quad\cdot
\bigl[(1-\lambda_c)+\lambda_c r_f^{\mathrm{temp}}(\mathbf p)\bigr].
\end{aligned}
\label{eq:8}
\end{equation}
The weight $r_f^{\mathrm{temp}}\in[0.5,1]$ is smallest at $\widetilde D_f^{\mathrm{temp}}=0.5$ and largest at 0 or 1. We set $\lambda_g=\lambda_c=0.5$. The score $\psi_f$ combines event activity, appearance boundaries, and temporal-difference reliability. Below, $D_f$ denotes rendered depth, distinct from $D_f^{\mathrm{temp}}$.

\subsubsection{Depth constraints and 3D aggregation}

We jointly render optical-axis depth $D_f$ from static and dynamic Gaussians and retain only finite depth values greater than $10^{-4}$. Candidate pixels are back-projected through the camera intrinsics and transformed to world coordinates. No support is generated where valid depth is unavailable. We group candidates into voxels of side length $h_v$ and use their centroids $\mathbf x_v$. Within-voxel scores are averaged and normalized across voxels to obtain $\psi_v\in[0,1]$; mean temporal differences determine dynamic labels $\chi_v^{\mathrm{dyn}}$. Appendix \ref{app:projection} gives the transforms and aggregation. The final candidate set is
\begin{equation}
\mathcal P_e=\operatorname{Top}_{N_{\mathrm{sup}}}
\{(\mathbf x_v,\chi_v^{\mathrm{dyn}},\psi_v):n_v\ge n_{\min},\,\psi_v\ge\tau_{\psi}\}.
\label{eq:10}
\end{equation}
The operator $\operatorname{Top}_{N_{\mathrm{sup}}}$ retains at most $N_{\mathrm{sup}}$ candidates by score, subject to the quota on dynamic candidates. Here, $n_{\min}$ is the minimum number of candidates per voxel, and $\tau_{\psi}$ is the retention threshold for normalized voxel scores. Appendix \ref{app:parameters} summarizes the shared settings and reported parameter ranges and identifies parameters that vary by scene. Let $\bar D_v^{\mathrm{temp}}$ denote the mean temporal difference within a voxel. The dynamic label is $\chi_v^{\mathrm{dyn}}=\mathbb I[\bar D_v^{\mathrm{temp}}/\max_{v\prime}\bar D_{v\prime}^{\mathrm{temp}}>0.5]$, with a label of 0 when the denominator is zero. The count $n_v$ measures repeated support from candidate points, not the number of independent views. Only dynamically labeled candidates are initialized as $\mathcal G_{\mathrm{sup}}^d$. Support augmentation is performed once, after which the joint reconstruction objective continually optimizes the geometry, appearance, and opacity of the added Gaussians.

\subsection{Contribution-Confidence Structural Motion Graph (CCSMG)}
\label{sec:structure}

Support augmentation determines which primitives can absorb reconstruction errors, but leaves their motion to be learned. Newly added support does not automatically provide reliable motion references, and existing Gaussians are also unevenly constrained by occlusion, weak texture, and brief visibility. Propagating all of their displacements indiscriminately could spread errors into neighboring regions. This brings us to the second assumption: that reliability can be estimated as though photometric optimization had already converged.

USplat4D evaluates analytical variance after pretraining~\cite{ref24}. In ESC-DGR, the Gaussian set and supervision evolve within one training run, so a fixed assessment at a converged state is unavailable. We instead accumulate each Gaussian's rendering contribution, contribution-weighted residual, and residual dispersion as optimization proceeds. Historical visibility, RGB quality, and event support complement these statistics in selecting motion references, which we call Keys. The remaining dynamic Gaussians are Non-keys. CCSMG then propagates reliable Key displacements to weakly observed neighborhoods using confidence weights. We call this process CC-guided flow. Here, CC denotes contribution confidence, measured by the contribution reliability $U_i$ in Eq. \eqref{eq:12}. CCSMG uses the augmented Gaussian set directly, without external trackers or pretrained trajectories. Its additional cost consists of one auxiliary rendering pass per statistics update and the maintenance of Keys and their neighborhoods.

\subsubsection{Rendering contribution statistics and confidence estimation}

A high contribution can reflect either accurate reconstruction or a footprint covering textureless regions. Mean error also hides local inconsistencies. We therefore measure contribution-weighted residuals and their dispersion in one auxiliary pass, keeping geometry and compositing order fixed. Exponential moving averages and valid-observation counts reduce sensitivity to occlusion and sampling jitter. The resulting confidence is a relative score based on observations, not a calibrated probability. It does not require photometric convergence.

Let the mean absolute RGB residual be $e(\mathbf p)=\tfrac13\|\widehat I(\mathbf p)-I(\mathbf p)\|_1$.

To compute accumulated contributions and the first two residual moments in one auxiliary rendering pass, we introduce an auxiliary attribute vector $\mathbf z_i=(z_{i,1},z_{i,2},z_{i,3})^{\mathsf T}$ for each Gaussian. With geometry, opacity, and compositing order fixed, we use these attributes as color inputs for differentiable rendering. The three output channels are denoted by $P_1$, $P_2$, and $P_3$. Let $\langle A,B\rangle=\sum_{\mathbf p}A(\mathbf p)B(\mathbf p)$. Then,
\begin{equation}
\begin{aligned}
\kappa_i&=\frac{\partial\langle P_1,\mathbf 1\rangle}{\partial z_{i,1}},\qquad
\bar e_i=\kappa_i^{-1}\frac{\partial\langle P_2,e\rangle}{\partial z_{i,2}},\\
m_{2,i}&=\kappa_i^{-1}\frac{\partial\langle P_3,e^2\rangle}{\partial z_{i,3}}.
\end{aligned}
\label{eq:contribution-statistics}
\end{equation}
Here, $\mathbf 1$ is an all-ones map, and $e^2$ denotes pixel-wise squaring. The quantity $\kappa_i$ is the accumulated rendering contribution over Gaussian $i$'s projected footprint. The quantities $\bar e_i$ and $m_{2,i}$ are the contribution-weighted mean residual and mean squared residual, respectively. The latter two are defined only when the Gaussian has a valid contribution.

The residual dispersion is $\upsilon_i=\sqrt{\max(m_{2,i}-\bar e_i^2,0)}$, giving the conservative error $\zeta_i=\operatorname{clip}(\bar e_i+0.25\upsilon_i,0,1)$. Contributions are normalized separately for static and dynamic sets using their valid median. We accumulate normalized contributions and errors into $C_i$ and $\bar\zeta_i$ by EMA, updating only on valid observations. Appendix \ref{app:statistics} gives the normalization, initialization, and update rules. With $o_i$ valid observations, contribution reliability is
\begin{equation}
U_i=\min(o_i/n_0,1)\,C_i\exp(-\bar\zeta_i/b_{\mathrm{err}}),
\label{eq:12}
\end{equation}
Here, $n_0$ controls the saturation scale of observation accumulation, and $b_{\mathrm{err}}$ controls residual decay. This measure reduces the influence of Gaussians with insufficient observations, weak contributions, or large footprint errors during structural propagation.

We further combine historical visibility $V_i$, RGB quality $\varrho_i$, and event support $\xi_i^{\mathrm{evt}}$ to obtain overall reliability:
\begin{equation}
Q_i = \beta_U U_i + \beta_V V_i + \beta_{\mathrm{rgb}} \varrho_i + \beta_{\mathrm{evt}} \xi_i^{\mathrm{evt}},
\label{eq:confidence}
\end{equation}
Here, $V_i$ measures the consistency of valid observations of a Gaussian over past contribution-statistics updates. The quantity $\varrho_i$ denotes historical RGB quality derived from rendering residuals in the projected region. The quantity $\xi_i^{\mathrm{evt}}$ denotes historical event support at the projected Gaussian location. All three are updated through exponential moving averages and normalized to $[0,1]$. Larger values indicate stronger observational evidence. The fusion weights $(\beta_U,\beta_V,\beta_{\mathrm{rgb}},\beta_{\mathrm{evt}})=(0.45,0.25,0.20,0.10)$ are fixed across scenes. Overall reliability $Q_i$ is used to filter and rank candidate motion references and to compute neighborhood propagation weights.

\subsubsection{Reliable Key selection and local neighborhoods}

USplat4D selects $2\%$ of Gaussians by uncertainty and assigns each Non-key the temporally nearest Key's edge set~\cite{ref24}, making guidance dependent on one Key. We instead filter candidates by contribution, reliability, and observation count, distribute them using voxel quotas, and combine several nearby Keys. Canonical-space connectivity remains fixed across rendered frames, reducing sensitivity to individual Key errors and changing neighborhoods.

Candidates must satisfy $C_i\ge\tau_C$, $Q_i\ge\tau_Q$, and $o_i\ge o_{\min}$. They are ranked by $C_i+0.25Q_i+0.02\min(o_i,5)$. Selection first retains the best candidate per voxel, then fills the remaining slots by score. The Key budget is proportional to the size of the dynamic set, subject to minimum and maximum limits. If too few candidates are available, selection falls back to all Gaussians using the same ranking. These coefficients are empirical, not probabilistically calibrated. Appendix \ref{app:anchors} gives the full selection procedure. For each Non-key $i$, $\mathcal N_i$ contains its $n_{\mathrm{nb}}$ nearest Keys in canonical space, at distances $d_{ij}=\|\boldsymbol\mu_i^0-\boldsymbol\mu_j^0\|_2$.
The joint distance and confidence weight is
\begin{equation}
\widetilde w_{ij}=
\exp\!\left(-\frac{d_{ij}^2}{2h_{\mathrm{graph}}^2}\right)
(C_j+\varepsilon_C)
[\varepsilon_Q+(1-\varepsilon_Q)Q_j],
\label{eq:15}
\end{equation}
Here, $h_{\mathrm{graph}}$ controls spatial decay. The lower bounds $\varepsilon_C>0$ and $\varepsilon_Q\in(0,1)$ apply to contribution and reliability weights, respectively. They prevent weak observations from completely removing local connections. We normalize only edges within the maximum distance $d_{\max}$:
\begin{equation}
w_{ij}=
\frac{\widetilde w_{ij}\mathbb I[d_{ij}\le d_{\max}]}
{\sum_{l\in\mathcal N_i}\widetilde w_{il}\mathbb I[d_{il}\le d_{\max}]}.
\label{eq:16}
\end{equation}
Nearby Keys with reliable observations receive greater propagation weights. Graph guidance is disabled when no valid incident edge exists, avoiding forced alignment with distant Keys.

\subsubsection{Structural motion propagation and constraints}

We assume that the low-frequency field is largely accurate in observed regions and use the graph to correct weakly observed motion incrementally. Direct pose assignment by dual-quaternion blending transfers Key poses rigidly. In contrast, our guidance weights each neighborhood edge by distance and reliability. Its strength increases gradually during training, and it is disabled when no valid reference is available. Stop-gradient decouples the reference from the corrected displacement. Rotation and scale remain controlled by the low-frequency branch.

Let $\mathbf m_{j,f}$ and $\mathbf q_{j,f}$ denote the translation and unit quaternion of a Key at frame $f$. A unit quaternion compactly represents 3D rotation with four components. For $t\in[t_f,t_{f+1}]$, let $\theta=(t-t_f)/(t_{f+1}-t_f)$. We interpolate Key motion between adjacent RGB frames as follows:
\begin{equation}
\begin{aligned}
\mathbf m_j(t)&=(1-\theta)\mathbf m_{j,f}+\theta\mathbf m_{j,f+1},\\
\mathbf q_j(t)&=
\frac{(1-\theta)\mathbf q_{j,f}+\theta\mathbf q_{j,f+1}}
{\left\|(1-\theta)\mathbf q_{j,f}+\theta\mathbf q_{j,f+1}\right\|_2}
\end{aligned}
\label{eq:key-temporal-interpolation}
\end{equation}
This interpolation provides translation and rotation states at continuous times. The quaternion combination is renormalized to satisfy the unit-length constraint and represent a valid rotation. Let $\mathbf r_{i,f}$ denote the effective local residual of a Non-key at frame $f$, including the $\gamma_r$ scaling of the learnable residual. Weighted local rigid transforms yield
\begin{equation}
\begin{aligned}
\boldsymbol\mu_{i,f}^{G}={}&
\sum_{j\in\mathcal N_i}w_{ij}
\big[\mathbf R(\mathbf q_{j,f})(\boldsymbol\mu_i^0-\boldsymbol\mu_j^0)\\
&+\boldsymbol\mu_j^0+\mathbf m_{j,f}\big]
+\mathbf r_{i,f},\\
\Delta\boldsymbol\mu_{i,f}^{G}={}&
\boldsymbol\mu_{i,f}^{G}-\boldsymbol\mu_i^0.
\end{aligned}
\label{eq:17}
\end{equation}
Here, $\mathbf R(\mathbf q_{j,f})$ is the rotation matrix corresponding to the quaternion. The parameter $\gamma_r$ controls local nonrigid freedom, and its scaling is already included in $\mathbf r_{i,f}$. Eq. \eqref{eq:17} linearly blends locally transformed positions. At times not included in training, $\mathbf m_j(t)$ and $\mathbf q_j(t)$ from Eq. \eqref{eq:key-temporal-interpolation} are used in the same propagation process. Key motion states are updated from low-frequency predictions at RGB training frames.

The graph reference softly guides base low-frequency motion through Eq. \eqref{eq:decomp}. For Non-key Gaussian $i$, guidance is applied only when its KNN neighborhood contains a Key within the valid distance and its historical event support, contribution reliability, and accumulated contribution jointly satisfy
\begin{equation}
\begin{aligned}
&\exists\,j\in\mathcal N_i:\ d_{ij}\leq d_{\max},
\qquad \xi_i^{\mathrm{evt}}\geq\tau_{\mathrm{evt}}^{G},\\
&U_i\geq\tau_U^{G},
\qquad C_i\geq\tau_C^{G},
\end{aligned}
\label{eq:structure-gate-conditions}
\end{equation}
When these conditions hold, $b_{i,f}^{G}=1$ and structural guidance is applied. Otherwise, $b_{i,f}^{G}=0$ and the base low-frequency motion is retained. Here, $\tau_{\mathrm{evt}}^{G}$, $\tau_U^{G}$, and $\tau_C^{G}$ are thresholds for historical event support, contribution reliability, and accumulated contribution, respectively. Key Gaussians do not receive neighborhood propagation and retain their own base low-frequency motion. This step uses accumulated observation statistics. Current event activity is used for local refinement in the next section. This separates stable structural references from immediate motion cues.

To constrain graph motion, we use
\begin{equation}
\begin{aligned}
\mathcal L_G={}&\lambda_a\mathcal L_{anchor}
+\lambda_n\mathcal L_{nonkey}
+\lambda_r\mathcal L_{rigid}\\
&+\lambda_t\mathcal L_{acc}
+\lambda_q\mathcal L_{rot}
+\lambda_{res}\mathcal L_{res}.
\end{aligned}
\label{eq:18}
\end{equation}
Here, $\mathcal L_{anchor}$ and $\mathcal L_{rot}$ anchor Key translation and rotation to detached low-frequency predictions, respectively. The term $\mathcal L_{nonkey}$ enforces consistency between low-frequency motion and graph references within valid neighborhoods. The term $\mathcal L_{rigid}$ preserves the distances between neighboring Keys, keeping their relative positions nearly fixed. These Key--Key links, which we call motion anchors, couple the Keys to a common structural reference that Non-keys can follow. The temporal term $\mathcal L_{acc}$ penalizes second-order differences of Key translation across adjacent frames. The term $\mathcal L_{res}$ limits Non-key residual magnitude. Together, these constraints allow local motion to share structural references while retaining limited nonrigid variation.

\subsection{Event-conditioned High-frequency Motion Refinement (EHMR)}
\label{sec:residual}

The preceding sections address which Gaussians can explain the observations and which motions can serve as references. Neither changes the temporal information available to frame-level deformation or the endpoint interpolation in Eq. \eqref{eq:key-temporal-interpolation}. Even a reliable pair of Key states may fail to describe rapid motion between them. The remaining edge displacements can therefore be averaged into blur and ghosting rather than represented explicitly.

Events provide observations at these intermediate times, but their sparsity and noise do not support independent recovery of a dense motion field. EHMR therefore refines the structurally guided motion instead of learning it again from events. The current Gaussian positions establish the correspondence with event features. Contribution reliability helps select the active set, and graph context conditions the predicted correction. Bounded position residuals then capture local deviations from the dominant motion, with zero residual for inactive Gaussians. Here, high frequency denotes these short-term local corrections, not a separate scene representation.

\subsubsection{Active Gaussian selection and event representation}

Event intensity alone cannot distinguish noise from motion and may miss occluded moving Gaussians. We combine event activity with reliability, accumulated contribution, and RGB reconstruction quality to select supported regions for refinement.

The normalized activity map $M_f\in[0,1]^{H_f\times W_f}$ describes current event response strength. We project the structurally guided low-frequency position onto the image plane to obtain $\mathbf p_{i,f}=\pi_f(\boldsymbol\mu_i^0+\widetilde{\Delta\boldsymbol\mu}_{i,f}^{L})$. The sampled activity at this location is $M_{i,f}=M_f(\mathbf p_{i,f})$.

Contribution statistics have largely stabilized when the high-frequency branch is enabled. We define a per-Gaussian reliability gate from contribution reliability $U_i$:
\begin{equation}
\rho_i^{\mathrm{hi}}=\rho_{\min}+(1-\rho_{\min})U_i,
\label{eq:high-reliability-gate}
\end{equation}
Here, $\rho_{\min}$ is the lower bound on the gate, and the superscript $\mathrm{hi}$ denotes the high-frequency residual branch. This mapping gives reliable Gaussians stronger responses without completely suppressing weakly observed Gaussians. It also ensures $\rho_i^{\mathrm{hi}}\in[\rho_{\min},1]$.

To reduce interference from isolated events and weakly observed Gaussians, we first construct a candidate set from current event activity and the reliability gate:
\begin{equation}
\mathcal J_f^{\mathrm{cand}}=\left\{i\mid M_{i,f}\rho_i^{\mathrm{hi}}>\tau_{\mathrm{act}}\right\},
\label{eq:active-candidates}
\end{equation}
Here, $\tau_{\mathrm{act}}$ is the activity candidate threshold. For $i\in\mathcal J_f^{\mathrm{cand}}$, we define an activity score combining local event consistency, accumulated contribution, and residual RGB reconstruction error:
\begin{equation}
\begin{aligned}
s_{i,f}^{\mathrm{act}}={}&M_{i,f}\rho_i^{\mathrm{hi}}
\,[\eta_h+(1-\eta_h)\chi_{i,f}^{\mathrm{evt}}]\\
&\cdot[\eta_C+(1-\eta_C)C_i]
\,[\eta_R+(1-\eta_R)(1-\varrho_i)],
\end{aligned}
\label{eq:active-score}
\end{equation}
Here, $\chi_{i,f}^{\mathrm{evt}}\in[0,1]$ measures event consistency in the projected neighborhood of Gaussian $i$ at frame $f$. The quantity $1-\varrho_i$ measures how poorly the RGB observations are reconstructed. The parameters $\eta_h$, $\eta_C$, and $\eta_R$ are lower weight bounds for the corresponding factors. Given the current frame's activity budget $B_f$, the active Gaussian set is
\begin{equation}
\mathcal J_f^{\mathrm{act}}=\operatorname{Top}_{B_f}
\left\{(i,s_{i,f}^{\mathrm{act}})\mid i\in\mathcal J_f^{\mathrm{cand}}\right\}.
\label{eq:active-set}
\end{equation}
Selection prioritizes edge Gaussians with sufficient event support. We partition screen space and allocate the budget according to the candidate count in each region to maintain spatial coverage of the active set.

For $i\in\mathcal J_f^{\mathrm{act}}$, we construct a feature map $\mathbf F_e\in\mathbb R^{H_f\times W_f\times16}$ from current events. This map encodes multiscale activity, polarity distributions, and local edges, yielding the per-Gaussian input
\begin{equation}
\mathbf f_{i,f}=
[\mathbf F_e(\mathbf p_{i,f}),M_{i,f},
\boldsymbol\mu_i^0,\widetilde{\Delta\boldsymbol\mu}_{i,f}^{L}].
\label{eq:19}
\end{equation}
Here, $\pi_f$ denotes camera projection, and square brackets indicate concatenation. By linking 3D positions to current events, the input conditions the residual prediction on observations aligned with the estimated low-frequency motion. Only Gaussians in $\mathcal J_f^{\mathrm{act}}$ undergo high-frequency residual prediction; all others retain their structurally guided low-frequency positions.

\subsubsection{Fusion of event and structural conditioning}

Current events indicate where the image changes, but do not by themselves specify how a Gaussian should move relative to the structural estimate. We therefore retain the graph-reference displacement and its difference from low-frequency motion as conditioning information. Residual prediction can thus use the reference constructed in Section \ref{sec:structure}, in addition to the reliability used for active-set selection. Global event statistics provide shared context, while per-Gaussian features preserve local activity and motion information. We inject this context through a gated residual after normalizing and projecting $\mathbf f_{i,f}$:
\begin{equation}
\begin{aligned}
\mathbf h_i^0&=\operatorname{GELU}(\mathbf W_x\operatorname{LN}(\mathbf f_{i,f})),\\
\mathbf c_i^{ctx}&=\mathbf W_e\mathbf e^{glob}+\mathbf W_u\mathbf f_{i,f},\\
\mathbf h_i^c&=\mathbf h_i^0+
\operatorname{Sigmoid}(\mathcal M_g([\mathbf h_i^0,\mathbf c_i^{ctx}]))
\odot\mathbf c_i^{ctx}.
\end{aligned}
\label{eq:20}
\end{equation}
Here, $\operatorname{LN}$ denotes layer normalization. The matrices $\mathbf W_x$, $\mathbf W_e$, and $\mathbf W_u$ are learnable projections, $\mathcal M_g$ is a gating network, and $\operatorname{Sigmoid}(\cdot)$ is the sigmoid function. The operator $\odot$ denotes element-wise multiplication. Each gate value lies in $(0,1)$ and controls the proportion of the corresponding context component injected into $\mathbf h_i^0$. Global statistics describe the current event distribution, while per-Gaussian inputs provide the corresponding activity and motion information.

The structural condition consists of the graph-reference displacement, the displacement difference between low-frequency motion and the graph reference, the reliability gate, local event consistency, and the rotation difference:
$\mathbf a_{i,f}=[\Delta\boldsymbol\mu_{i,f}^G,\widetilde{\Delta\boldsymbol\mu}_{i,f}^L-\Delta\boldsymbol\mu_{i,f}^G,\rho_i^{\mathrm{hi}},\chi_{i,f}^{\mathrm{evt}},\Delta\mathbf q_{i,f}^{ctx}]$.
Here, $\chi_{i,f}^{\mathrm{evt}}$ is the local event consistency defined in Eq. \eqref{eq:active-score}. The quantity $\Delta\mathbf q_{i,f}^{ctx}$ is the difference between the normalized low-frequency quaternion and graph-reference quaternion at the current frame. Independent linear projections map the structural condition $\mathbf a_{i,f}$ to a structural key and value. The feature $\mathbf h_i^c$ is also projected into a query. A sigmoid maps their per-Gaussian similarity to a gating weight, which controls residual injection of the structural value into $\mathbf h_i^c$ to produce the fused feature $\widetilde{\mathbf h}_{i,f}$. Motion states are detached when constructing structural conditions. The fused feature then passes through one KAN residual block and two MLP residual blocks:
\begin{equation}
\mathbf h_{i,f}^{out}=
\mathcal R_2^{\mathrm{MLP}}\!\left(
\mathcal R_1^{\mathrm{MLP}}\!\left(
\mathcal R^{\mathrm{KAN}}(\widetilde{\mathbf h}_{i,f})
\right)\right).
\label{eq:21}
\end{equation}
Here, $\mathcal R^{\mathrm{KAN}}$ is a residual block with two KAN layers~\cite{ref53}. Each edge function combines a learnable expansion in zero-order, piecewise-constant basis functions with a SiLU base branch. The piecewise basis approximates local nonlinearities in the residual function, while $\mathcal R_1^{\mathrm{MLP}}$ and $\mathcal R_2^{\mathrm{MLP}}$ further combine the conditioning features.

\subsubsection{Bounded residuals and motion composition}

Unconstrained residuals may absorb motion that should be modeled by the low-frequency field and accumulate into temporal drift. We therefore use the Gaussian's minimum axial scale to define a local radius that bounds each correction. After applying $\tanh$ and clipping, we add the residual to the low-frequency displacement. Inactive Gaussians receive zero residuals. The branch predicts only 3D position changes. Its output is controlled jointly by a gating score $g_{i,f}$ and a scalar $a_{\mathrm{amp}}$:
\begin{equation}
\begin{aligned}
g_{i,f}&=\mathbf w_g^{\mathsf T}\mathbf h_{i,f}^{out}+b_g,\\
\Delta\boldsymbol\mu_{i,f}^{\mathrm{hi}}
&=|a_{\mathrm{amp}}|\,\operatorname{Sigmoid}(g_{i,f})\,
\mathbf W_{\mu}\mathbf h_{i,f}^{out}.
\end{aligned}
\label{eq:22}
\end{equation}
Here, $\mathbf w_g$ is the learnable weight vector of the gate output layer, and $b_g$ is a learnable scalar bias. Together, they map $\mathbf h_{i,f}^{out}$ to the gating score $g_{i,f}$. The scalar $a_{\mathrm{amp}}$ is a learnable amplitude, and $\mathbf W_\mu$ is the position mapping. Zero initialization of this mapping preserves the existing motion when the branch is first introduced.

To keep refinement local, we use the Gaussian's minimum axial scale $s_i^{\min}$.

\noindent The radius is $\ell_i=\operatorname{clip}(\gamma_s s_i^{\min},\ell_{\min},\ell_{\max})$, where $\gamma_s=0.75$, $\ell_{\min}=0.001$, and $\ell_{\max}=0.05$. The latter two are expressed in the same units as the canonical-space coordinates.
The actual high-frequency compensation and its composition with low-frequency motion are
\begin{equation}
\begin{aligned}
\widehat{\boldsymbol\delta}_{i,f}^{\mathrm{hi}}
&=\rho_i^{\mathrm{hi}} M_{i,f}
[\eta_h+(1-\eta_h)\chi_{i,f}^{\mathrm{evt}}]\\
&\quad\cdot
\operatorname{clip}_{\|\cdot\|_2\le\ell_i}
\!\left[s_{\mathrm{hi}}(k)\tanh\!\left(\Delta\boldsymbol\mu_{i,f}^{\mathrm{hi}}\right)\right],\\
\boldsymbol\delta_{i,f}^{\mathrm{hi}}
&=
\begin{cases}
\widehat{\boldsymbol\delta}_{i,f}^{\mathrm{hi}}, & i\in\mathcal J_f^{\mathrm{act}},\\
\mathbf 0, & i\notin\mathcal J_f^{\mathrm{act}},
\end{cases}\\[5pt]
\Delta\boldsymbol\mu_{i,f}={}&
\widetilde{\Delta\boldsymbol\mu}_{i,f}^L
+\boldsymbol\delta_{i,f}^{\mathrm{hi}}.
\end{aligned}
\label{eq:23}
\end{equation}
Here, $s_{\mathrm{hi}}(k)$ is a progressive scaling coefficient. The parameter $\eta_h\in[0,1]$ is the lower bound on the event-consistency weight, as defined in Eq. \eqref{eq:active-score}. Thus, $\eta_h+(1-\eta_h)\chi_{i,f}^{\mathrm{evt}}$ preserves base compensation while increasing with local event consistency. The vector $\widehat{\boldsymbol\delta}_{i,f}^{\mathrm{hi}}$ denotes the candidate correction weighted by event evidence and reliability. Vector clipping preserves its direction and limits its norm to $\ell_i$. For Gaussians outside the active set, the piecewise definition directly sets $\boldsymbol\delta_{i,f}^{\mathrm{hi}}=\mathbf 0$. Since activity and reliability factors do not exceed 1, the actual compensation satisfies $\|\boldsymbol\delta_{i,f}^{\mathrm{hi}}\|_2\le\ell_i$. This constraint limits only the local residual, not the magnitude of the main low-frequency motion.


\subsection{Joint objective and optimization strategy}
\label{sec:objective}

The three modules receive joint RGB and event supervision through the same dynamic representation. The total objective consists of reconstruction terms, graph constraints, and four auxiliary regularizers:
\begin{equation}
\begin{aligned}
\mathcal L={}&\mathcal L_{\mathrm{rgb}}
+\lambda_{\mathrm{evt}}(k)\mathcal L_{\mathrm{evt}}
+\lambda_{\mathrm{th}}\mathcal L_{\mathrm{th}}\\
&+\lambda_G(k)\mathcal L_G
+\lambda_d\mathcal L_{dyn}
+\lambda_b\mathcal L_{edge}\\
&+\lambda_{art}\mathcal L_{art}
+\lambda_{mag}(k)\mathcal L_{mag}.
\end{aligned}
\label{eq:26}
\end{equation}
Here, $\mathcal L_{\mathrm{rgb}}$ combines $\ell_1$ and D-SSIM for RGB reconstruction. The term $\mathcal L_{\mathrm{evt}}$ is the event residual in log intensity, and $\mathcal L_{\mathrm{th}}$ is the threshold calibration term~\cite{ref14}. The graph term $\mathcal L_G$ is defined in Eq. \eqref{eq:18}. The reconstruction terms alone do not directly address four sources of bias in joint training. Static backgrounds dominate pixel averages, and dynamic boundaries are susceptible to smoothing. Density control may retain Gaussians that absorb errors without contributing to image formation. Unconstrained residual compensation may also become unbounded. The four auxiliary regularizers address these issues, respectively.
Specifically, $\mathcal L_{dyn}$ applies a Charbonnier RGB penalty in event-active regions. The term $\mathcal L_{edge}$ matches image gradients in locally consistent event regions. The term $\mathcal L_{art}$ penalizes the opacity of persistently high-error, low-contribution Gaussians, and $\mathcal L_{mag}$ penalizes active residual magnitude. Appendix \ref{app:regularizers} gives their exact definitions. Appendix \ref{app:parameters} summarizes shared loss weights and identifies scene-dependent weights. The training phases are described below.

Clipping in Eq. \eqref{eq:23} enforces displacement bounds, while the regularizers favor reconstruction quality within those bounds. Because the corrections are bounded, well-reconstructed backgrounds are largely preserved. The training schedule follows the dependencies between support and motion: static warm-up provides the geometry used for depth-constrained support injection, after which new and existing Gaussians undergo joint optimization and density control. Graph guidance begins after densification ends, allowing motion references to be established on the resulting set. The residual branch is introduced late, when contribution statistics and dominant motion have become more stable. Guidance and residual coefficients increase progressively so that later corrections do not amplify unstable estimates. These are phases of one optimization run, not separate pretrained scene models.

\section{Experiments}

\subsection{Experimental settings}
\label{sec:experiment-settings}

\textbf{Datasets and baselines.} We evaluate on eight synthetic and four real scenes from the E-D3DGS benchmark~\cite{ref14}. The synthetic scenes are rendered in Blender with a camera completing a full orbit around the object, and they include fast motion and large displacements. The real scenes are captured using a synchronized turntable and beam-splitter system and cover mechanical motion, vehicle motion, flower deformation, and bird motion, enabling evaluation under real imaging conditions.

We compare against 3DGS, TiNeuVox, K-Planes, 4DGS, Deformable-3DGS, E-D3DGS, and FastEventDGS~\cite{ref42}. For real scenes, we compare against 4DGS, Deformable-3DGS, E-D3DGS, and FastEventDGS. Alongside the direct baseline E-D3DGS, we include the recent FastEventDGS method to broaden the comparison with event-based dynamic Gaussian reconstruction. We retain its original input organization and training pipeline, add RGB reconstruction supervision to its objective, and use the same initialization, camera poses, and event windows as our method.

DEGS, E-4DGS, STD-GS, and Event-4DGS are excluded because executable implementations or complete reproduction settings were unavailable at the time of our check (November 10, 2026). Appendix \ref{app:baselines} details the method-specific restrictions and distinguishes E-4DGS from Event-4DGS. Our performance conclusions cover only the evaluated methods; previously reported comparisons are not direct evidence under our protocol.

All compared methods are retrained and evaluated using the same frame split: 30 training frames per scene and 29 held-out intermediate frames for testing. The number of training iterations, initial point clouds, and evaluation code are kept consistent to compare motion modeling across representations under a common protocol. The additional FastEventDGS results include full-image PSNR, SSIM, LPIPS, and FPS for all twelve scenes, but do not yet include dynamic-region metrics; comparisons with this method are therefore restricted to full-image quality. Ablations use the same training data, test frames, and evaluation code to assess the added modules and the overall structural graph design.

\textbf{Evaluation metrics.} Because our method targets support and motion modeling in dynamic regions, we use dPSNR, dSSIM, and dLPIPS as the primary metrics of pixel reconstruction, structural similarity, and perceptual distance, respectively. Full-image PSNR, SSIM, and LPIPS~\cite{ref33,ref34} provide complementary measures of overall reconstruction quality. Dynamic regions are identified using the E-D3DGS procedure. The E-D3DGS baseline, full model, and all ablation variants share the same region-generation function, thresholds, crop size, and evaluation implementation to avoid bias from manual selection or method-specific settings. Rendering efficiency is measured in frames per second (FPS), with higher values indicating faster rendering. All experiments use the same hardware (NVIDIA RTX PRO 6000), evaluation resolution, and evaluation code.

\textbf{Implementation details.} ESC-DGR and E-D3DGS share the codebase, inputs, initialization, warm-up, decomposition parameters, hardware, and seed settings, and are trained for 50,000 iterations. ESC-DGR adds support at iteration 3,001, enables event supervision at 8,000, ends densification and introduces graph guidance at 18,000, and incorporates residual refinement from 35,000. Threshold pretraining takes 7,000 steps. Quality metrics are reported as means and standard deviations over three independent runs with different random seeds; E-D3DGS, ESC-DGR, and all ablation variants use the same three seeds. Cross-scene results use arithmetic means. FPS and dynamic Gaussian counts are taken from a single run per scene and averaged over scenes. Appendix \ref{app:training} gives reporting conventions and network settings, and Appendix \ref{app:parameters} lists parameters.

\textbf{Test protocol.} We evaluate joint view--time interpolation along the moving monocular capture trajectory. ESC-DGR constructs features, active masks, and residuals using the trained Gaussians, the target camera and time, and an aligned event window $\mathcal E_{\mathrm{obs}}(t^\ast)$ from the same calibrated camera. Both ESC-DGR and E-D3DGS train on identical RGB and event observations with the same preprocessing. All events used for test-time conditioning belong to this shared training stream, so no additional capture is required. Training RGB-error statistics are frozen at test time, and test RGB images are used only for evaluation. Nevertheless, the inference inputs differ: ESC-DGR reads target-time events, whereas E-D3DGS does not. The variant without EHMR matches the baseline's inference inputs and still improves mean dPSNR across twelve scenes by 0.83~dB (23.60 versus 22.77).

\begin{table*}[width=\textwidth,pos=t]
  \centering
  \caption{Full-image reconstruction quality and rendering speed on the synthetic dynamic benchmark. Quality metrics are reported as means and standard deviations. The small $\pm$ values shown at the lower right of each mean denote standard deviations. Leading zeros are retained for means below 1. The best and second-best means are marked in bold and underlined, respectively.}
  \label{tab:2}
  \renewcommand{\metricstdcorner}[2]{#1\kern0.08em\raisebox{-0.45ex}{\normalfont\fontsize{5.2}{5.2}\selectfont \ensuremath{\pm}#2}}
  \rmfamily
  \fontsize{8.5}{10}\selectfont
  \setlength{\tabcolsep}{3pt}
  \setlength{\heavyrulewidth}{.5pt}
  \setlength{\lightrulewidth}{.3pt}
  \setlength{\cmidrulewidth}{.3pt}
  \setlength{\arrayrulewidth}{.3pt}
  \renewcommand{\arraystretch}{1.12}
  \begin{tabularx}{\textwidth}{@{}c|>{\hsize=1.08\hsize\linewidth=\hsize}Y Y Y >{\hsize=.92\hsize\linewidth=\hsize}Y|>{\hsize=1.08\hsize\linewidth=\hsize}Y Y Y >{\hsize=.92\hsize\linewidth=\hsize}Y}
    \toprule
    \multirow{2}{*}{Method} & \multicolumn{4}{c|}{Lego} & \multicolumn{4}{c}{Hotdog} \\
    \cmidrule(l{4.7pt}r{5pt}){2-5}\cmidrule(l{4.85pt}r{5pt}){6-9}
    & PSNR$\uparrow$ & SSIM$\uparrow$ & LPIPS$\downarrow$ & FPS$\uparrow$ & PSNR$\uparrow$ & SSIM$\uparrow$ & LPIPS$\downarrow$ & FPS$\uparrow$ \\
    \midrule
    3DGS~\cite{ref1} & \metricstdcorner{23.55}{0.05} & \metricstdcorner{0.9172}{0.0004} & \metricstdcorner{0.0868}{0.0006} & \textbf{314} & \metricstdcorner{30.11}{0.04} & \metricstdcorner{0.9524}{0.0001} & \metricstdcorner{0.0640}{0.0004} & \textbf{304} \\
    TiNeuVox~\cite{ref15} & \metricstdcorner{22.19}{0.05} & \metricstdcorner{0.9155}{0.0004} & \metricstdcorner{0.0691}{0.0005} & 0.67 & \metricstdcorner{30.93}{0.04} & \metricstdcorner{0.9552}{0.0001} & \metricstdcorner{0.0304}{0.0002} & 0.56 \\
    K-Planes~\cite{ref4} & \metricstdcorner{23.38}{0.04} & \metricstdcorner{0.9123}{0.0005} & \metricstdcorner{0.0553}{0.0007} & 2.77 & \metricstdcorner{26.95}{0.05} & \metricstdcorner{0.9465}{0.0002} & \metricstdcorner{0.0412}{0.0003} & 3.42 \\
    4DGS~\cite{ref6} & \metricstdcorner{26.17}{0.06} & \metricstdcorner{0.9282}{0.0005} & \metricstdcorner{0.0674}{0.0011} & 125 & \metricstdcorner{33.93}{0.06} & \metricstdcorner{0.9667}{0.0002} & \metricstdcorner{0.0166}{0.0001} & 189 \\
    Deformable-3DGS~\cite{ref7} & \metricstdcorner{23.81}{0.04} & \metricstdcorner{0.9230}{0.0003} & \metricstdcorner{0.0524}{0.0005} & 113 & \metricstdcorner{32.99}{0.04} & \metricstdcorner{0.9608}{0.0001} & \metricstdcorner{0.0175}{0.0001} & 193 \\
    E-D3DGS~\cite{ref14} & \metricstdcorner{\underline{31.81}}{0.10} & \metricstdcorner{\underline{0.9661}}{0.0009} & \metricstdcorner{\underline{0.0183}}{0.0004} & \underline{237} & \metricstdcorner{\underline{36.15}}{0.02} & \metricstdcorner{\underline{0.9733}}{0.0002} & \metricstdcorner{\underline{0.0131}}{0.0001} & \underline{282} \\
    FastEventDGS~\cite{ref42} & \metricstdcorner{24.13}{0.08} & \metricstdcorner{0.9071}{0.0012} & \metricstdcorner{0.0773}{0.0004} & 175 & \metricstdcorner{25.64}{0.07} & \metricstdcorner{0.9323}{0.0002} & \metricstdcorner{0.0955}{0.0002} & 193 \\
    ESC-DGR & \metricstdcorner{\textbf{32.29}}{0.07} & \metricstdcorner{\textbf{0.9716}}{0.0007} & \metricstdcorner{\textbf{0.0152}}{0.0003} & 196 & \metricstdcorner{\textbf{36.42}}{0.09} & \metricstdcorner{\textbf{0.9762}}{0.0002} & \metricstdcorner{\textbf{0.0126}}{0.0001} & 235 \\
    \midrule
    \multirow{2}{*}{Method} & \multicolumn{4}{c|}{Materials} & \multicolumn{4}{c}{Musicbox} \\
    \cmidrule(l{4.7pt}r{5pt}){2-5}\cmidrule(l{4.85pt}r{5pt}){6-9}
    & PSNR$\uparrow$ & SSIM$\uparrow$ & LPIPS$\downarrow$ & FPS$\uparrow$ & PSNR$\uparrow$ & SSIM$\uparrow$ & LPIPS$\downarrow$ & FPS$\uparrow$ \\
    \midrule
    3DGS~\cite{ref1} & \metricstdcorner{28.14}{0.04} & \metricstdcorner{0.9683}{0.0002} & \metricstdcorner{0.0595}{0.0005} & \textbf{347} & \metricstdcorner{19.20}{0.05} & \metricstdcorner{0.9552}{0.0005} & \metricstdcorner{0.0616}{0.0007} & \textbf{316} \\
    TiNeuVox~\cite{ref15} & \metricstdcorner{26.31}{0.05} & \metricstdcorner{0.9428}{0.0002} & \metricstdcorner{0.0578}{0.0004} & 0.64 & \metricstdcorner{18.89}{0.05} & \metricstdcorner{0.8174}{0.0006} & \metricstdcorner{0.1627}{0.0015} & 0.79 \\
    K-Planes~\cite{ref4} & \metricstdcorner{31.09}{0.08} & \metricstdcorner{0.9783}{0.0003} & \metricstdcorner{0.0092}{0.0001} & 3.14 & \metricstdcorner{20.51}{0.06} & \metricstdcorner{0.8438}{0.0008} & \metricstdcorner{0.0983}{0.0014} & 3.33 \\
    4DGS~\cite{ref6} & \metricstdcorner{29.36}{0.07} & \metricstdcorner{0.9778}{0.0004} & \metricstdcorner{0.0610}{0.0006} & 146 & \metricstdcorner{23.23}{0.07} & \metricstdcorner{0.9316}{0.0005} & \metricstdcorner{0.0445}{0.0006} & 91 \\
    Deformable-3DGS~\cite{ref7} & \metricstdcorner{33.12}{0.05} & \metricstdcorner{0.9823}{0.0002} & \metricstdcorner{0.0044}{0.0001} & 142 & \metricstdcorner{21.98}{0.06} & \metricstdcorner{0.9225}{0.0005} & \metricstdcorner{0.0550}{0.0005} & 60 \\
    E-D3DGS~\cite{ref14} & \metricstdcorner{\underline{38.02}}{0.08} & \metricstdcorner{\underline{0.9894}}{0.0004} & \metricstdcorner{\underline{0.0031}}{0.0001} & \underline{332} & \metricstdcorner{\underline{30.74}}{0.08} & \metricstdcorner{\underline{0.9587}}{0.0009} & \metricstdcorner{\underline{0.0292}}{0.0006} & \underline{119} \\
    FastEventDGS~\cite{ref42} & \metricstdcorner{30.22}{0.06} & \metricstdcorner{0.9237}{0.0003} & \metricstdcorner{0.0237}{0.0003} & 172 & \metricstdcorner{21.47}{0.05} & \metricstdcorner{0.9048}{0.0007} & \metricstdcorner{0.1138}{0.0012} & 71 \\
    ESC-DGR & \metricstdcorner{\textbf{38.47}}{0.08} & \metricstdcorner{\textbf{0.9953}}{0.0004} & \metricstdcorner{\textbf{0.0024}}{0.0001} & 257 & \metricstdcorner{\textbf{31.19}}{0.08} & \metricstdcorner{\textbf{0.9665}}{0.0010} & \metricstdcorner{\textbf{0.0238}}{0.0007} & 89 \\
    \midrule
    \multirow{2}{*}{Method} & \multicolumn{4}{c|}{Celestial globe} & \multicolumn{4}{c}{Fan} \\
    \cmidrule(l{4.7pt}r{5pt}){2-5}\cmidrule(l{4.85pt}r{5pt}){6-9}
    & PSNR$\uparrow$ & SSIM$\uparrow$ & LPIPS$\downarrow$ & FPS$\uparrow$ & PSNR$\uparrow$ & SSIM$\uparrow$ & LPIPS$\downarrow$ & FPS$\uparrow$ \\
    \midrule
    3DGS~\cite{ref1} & \metricstdcorner{18.74}{0.04} & \metricstdcorner{0.9213}{0.0003} & \metricstdcorner{0.1213}{0.0010} & \textbf{262} & \metricstdcorner{21.26}{0.04} & \metricstdcorner{0.8915}{0.0001} & \metricstdcorner{0.1181}{0.0006} & \textbf{378} \\
    TiNeuVox~\cite{ref15} & \metricstdcorner{13.49}{0.04} & \metricstdcorner{0.7324}{0.0006} & \metricstdcorner{0.2886}{0.0032} & 0.83 & \metricstdcorner{19.85}{0.05} & \metricstdcorner{0.8762}{0.0002} & \metricstdcorner{0.1087}{0.0022} & 0.63 \\
    K-Planes~\cite{ref4} & \metricstdcorner{16.38}{0.07} & \metricstdcorner{0.8583}{0.0006} & \metricstdcorner{0.0762}{0.0008} & 3.07 & \metricstdcorner{20.09}{0.05} & \metricstdcorner{0.8857}{0.0002} & \metricstdcorner{0.0953}{0.0010} & 3.00 \\
    4DGS~\cite{ref6} & \metricstdcorner{21.57}{0.05} & \metricstdcorner{0.9435}{0.0004} & \metricstdcorner{0.0621}{0.0006} & 73 & \metricstdcorner{26.02}{0.04} & \metricstdcorner{0.9433}{0.0002} & \metricstdcorner{0.0355}{0.0003} & 156 \\
    Deformable-3DGS~\cite{ref7} & \metricstdcorner{22.94}{0.04} & \metricstdcorner{0.9612}{0.0002} & \metricstdcorner{0.0360}{0.0002} & 57 & \metricstdcorner{24.44}{0.04} & \metricstdcorner{0.9275}{0.0001} & \metricstdcorner{0.0534}{0.0002} & 129 \\
    E-D3DGS~\cite{ref14} & \metricstdcorner{\underline{28.81}}{0.05} & \metricstdcorner{\underline{0.9744}}{0.0004} & \metricstdcorner{\underline{0.0212}}{0.0002} & \underline{94} & \metricstdcorner{\underline{30.11}}{0.05} & \metricstdcorner{\underline{0.9638}}{0.0001} & \metricstdcorner{\underline{0.0257}}{0.0002} & \underline{217} \\
    FastEventDGS~\cite{ref42} & \metricstdcorner{25.37}{0.06} & \metricstdcorner{0.9659}{0.0002} & \metricstdcorner{0.0384}{0.0003} & 65 & \metricstdcorner{21.73}{0.03} & \metricstdcorner{0.9139}{0.0002} & \metricstdcorner{0.1101}{0.0016} & 146 \\
    ESC-DGR & \metricstdcorner{\textbf{29.10}}{0.05} & \metricstdcorner{\textbf{0.9782}}{0.0003} & \metricstdcorner{\textbf{0.0192}}{0.0002} & 82 & \metricstdcorner{\textbf{30.34}}{0.04} & \metricstdcorner{\textbf{0.9655}}{0.0001} & \metricstdcorner{\textbf{0.0220}}{0.0002} & 163 \\
    \midrule
    \multirow{2}{*}{Method} & \multicolumn{4}{c|}{Waterwheel} & \multicolumn{4}{c}{Man} \\
    \cmidrule(l{4.7pt}r{5pt}){2-5}\cmidrule(l{4.85pt}r{5pt}){6-9}
    & PSNR$\uparrow$ & SSIM$\uparrow$ & LPIPS$\downarrow$ & FPS$\uparrow$ & PSNR$\uparrow$ & SSIM$\uparrow$ & LPIPS$\downarrow$ & FPS$\uparrow$ \\
    \midrule
    3DGS~\cite{ref1} & \metricstdcorner{19.23}{0.03} & \metricstdcorner{0.8835}{0.0002} & \metricstdcorner{0.1121}{0.0008} & \textbf{301} & \metricstdcorner{20.88}{0.03} & \metricstdcorner{0.8746}{0.0005} & \metricstdcorner{0.1173}{0.0019} & \textbf{275} \\
    TiNeuVox~\cite{ref15} & \metricstdcorner{17.80}{0.06} & \metricstdcorner{0.7701}{0.0003} & \metricstdcorner{0.2467}{0.0027} & 0.67 & \metricstdcorner{22.74}{0.05} & \metricstdcorner{0.8853}{0.0006} & \metricstdcorner{0.0731}{0.0009} & 0.59 \\
    K-Planes~\cite{ref4} & \metricstdcorner{20.79}{0.06} & \metricstdcorner{0.9056}{0.0003} & \metricstdcorner{0.0561}{0.0005} & 2.98 & \metricstdcorner{21.39}{0.06} & \metricstdcorner{0.8592}{0.0006} & \metricstdcorner{0.0797}{0.0011} & 2.94 \\
    4DGS~\cite{ref6} & \metricstdcorner{21.35}{0.06} & \metricstdcorner{0.9044}{0.0004} & \metricstdcorner{0.0610}{0.0007} & 98 & \metricstdcorner{23.81}{0.05} & \metricstdcorner{0.9127}{0.0005} & \metricstdcorner{0.0732}{0.0009} & 91 \\
    Deformable-3DGS~\cite{ref7} & \metricstdcorner{20.18}{0.03} & \metricstdcorner{0.9048}{0.0002} & \metricstdcorner{0.0554}{0.0004} & 73 & \metricstdcorner{23.06}{0.03} & \metricstdcorner{0.9055}{0.0002} & \metricstdcorner{0.0520}{0.0004} & 57 \\
    E-D3DGS~\cite{ref14} & \metricstdcorner{\underline{28.30}}{0.06} & \metricstdcorner{\underline{0.9485}}{0.0003} & \metricstdcorner{\underline{0.0343}}{0.0003} & \underline{154} & \metricstdcorner{\underline{28.01}}{0.06} & \metricstdcorner{\underline{0.9420}}{0.0005} & \metricstdcorner{\underline{0.0353}}{0.0005} & \underline{179} \\
    FastEventDGS~\cite{ref42} & \metricstdcorner{21.39}{0.05} & \metricstdcorner{0.8919}{0.0003} & \metricstdcorner{0.1151}{0.0006} & 106 & \metricstdcorner{25.35}{0.04} & \metricstdcorner{0.9357}{0.0006} & \metricstdcorner{0.0470}{0.0006} & 128 \\
    ESC-DGR & \metricstdcorner{\textbf{28.72}}{0.04} & \metricstdcorner{\textbf{0.9512}}{0.0002} & \metricstdcorner{\textbf{0.0322}}{0.0003} & 114 & \metricstdcorner{\textbf{28.76}}{0.05} & \metricstdcorner{\textbf{0.9526}}{0.0005} & \metricstdcorner{\textbf{0.0302}}{0.0003} & 136 \\
    \bottomrule
  \end{tabularx}
\end{table*}

\begin{table*}[width=\textwidth,pos=t]
  \centering
  \caption{Full-image reconstruction quality and rendering speed on four real scenes. Quality metrics are reported as means and standard deviations. The small $\pm$ values shown at the lower right of each mean denote standard deviations. Leading zeros are retained for means below 1. The best and second-best means are marked in bold and underlined, respectively.}
  \label{tab:real-world}
  \renewcommand{\metricstdcorner}[2]{#1\kern0.08em\raisebox{-0.45ex}{\normalfont\fontsize{5.2}{5.2}\selectfont \ensuremath{\pm}#2}}
  \rmfamily
  \fontsize{8.5}{10}\selectfont
  \setlength{\tabcolsep}{3pt}
  \renewcommand{\arraystretch}{1.12}
  \begin{tabularx}{\textwidth}{@{}c|>{\hsize=1.08\hsize\linewidth=\hsize}Y Y Y >{\hsize=.92\hsize\linewidth=\hsize}Y|>{\hsize=1.08\hsize\linewidth=\hsize}Y Y Y >{\hsize=.92\hsize\linewidth=\hsize}Y}
    \toprule
    \multirow{2}{*}{Method} & \multicolumn{4}{c|}{Excavator} & \multicolumn{4}{c}{Jeep} \\
    \cmidrule(l{4.7pt}r{5pt}){2-5}\cmidrule(l{4.85pt}r{5pt}){6-9}
    & PSNR$\uparrow$ & SSIM$\uparrow$ & LPIPS$\downarrow$ & FPS$\uparrow$ & PSNR$\uparrow$ & SSIM$\uparrow$ & LPIPS$\downarrow$ & FPS$\uparrow$ \\
    \midrule
    4DGS~\cite{ref6} & \metricstdcorner{28.46}{0.06} & \metricstdcorner{0.9151}{0.0015} & \metricstdcorner{0.1125}{0.0032} & 164 & \metricstdcorner{26.53}{0.08} & \metricstdcorner{0.8543}{0.0022} & \metricstdcorner{0.0755}{0.0031} & 83 \\
    Deformable-3DGS~\cite{ref7} & \metricstdcorner{25.89}{0.04} & \metricstdcorner{0.9123}{0.0009} & \metricstdcorner{0.1178}{0.0018} & 109 & \metricstdcorner{25.49}{0.05} & \metricstdcorner{0.8426}{0.0012} & \metricstdcorner{0.0902}{0.0008} & 94 \\
    E-D3DGS~\cite{ref14} & \metricstdcorner{\underline{31.13}}{0.06} & \metricstdcorner{\underline{0.9217}}{0.0017} & \metricstdcorner{\underline{0.0782}}{0.0024} & \textbf{220} & \metricstdcorner{\underline{30.76}}{0.10} & \metricstdcorner{\underline{0.9052}}{0.0020} & \metricstdcorner{\underline{0.0636}}{0.0009} & \underline{126} \\
    FastEventDGS~\cite{ref42} & \metricstdcorner{26.74}{0.05} & \metricstdcorner{0.9156}{0.0012} & \metricstdcorner{0.1162}{0.0013} & 157 & \metricstdcorner{26.24}{0.08} & \metricstdcorner{0.8587}{0.0016} & \metricstdcorner{0.2391}{0.0017} & \textbf{150} \\
    ESC-DGR & \metricstdcorner{\textbf{31.67}}{0.09} & \metricstdcorner{\textbf{0.9308}}{0.0016} & \metricstdcorner{\textbf{0.0641}}{0.0019} & \underline{170} & \metricstdcorner{\textbf{31.39}}{0.08} & \metricstdcorner{\textbf{0.9202}}{0.0018} & \metricstdcorner{\textbf{0.0572}}{0.0009} & 99 \\
    \midrule
    \multirow{2}{*}{Method} & \multicolumn{4}{c|}{Flowers} & \multicolumn{4}{c}{Bird} \\
    \cmidrule(l{4.7pt}r{5pt}){2-5}\cmidrule(l{4.85pt}r{5pt}){6-9}
    & PSNR$\uparrow$ & SSIM$\uparrow$ & LPIPS$\downarrow$ & FPS$\uparrow$ & PSNR$\uparrow$ & SSIM$\uparrow$ & LPIPS$\downarrow$ & FPS$\uparrow$ \\
    \midrule
    4DGS~\cite{ref6} & \metricstdcorner{25.87}{0.04} & \metricstdcorner{0.8525}{0.0014} & \metricstdcorner{0.1183}{0.0030} & 96 & \metricstdcorner{26.33}{0.08} & \metricstdcorner{0.8808}{0.0007} & \metricstdcorner{0.1165}{0.0017} & 172 \\
    Deformable-3DGS~\cite{ref7} & \metricstdcorner{25.36}{0.04} & \metricstdcorner{0.8786}{0.0006} & \metricstdcorner{\underline{0.0752}}{0.0012} & 82 & \metricstdcorner{25.59}{0.05} & \metricstdcorner{0.8413}{0.0008} & \metricstdcorner{0.0897}{0.0007} & 97 \\
    E-D3DGS~\cite{ref14} & \metricstdcorner{\underline{28.53}}{0.04} & \metricstdcorner{\underline{0.9074}}{0.0015} & \metricstdcorner{0.0797}{0.0022} & \textbf{231} & \metricstdcorner{\underline{31.14}}{0.08} & \metricstdcorner{\underline{0.9182}}{0.0009} & \metricstdcorner{\underline{0.0816}}{0.0011} & \textbf{258} \\
    FastEventDGS~\cite{ref42} & \metricstdcorner{27.24}{0.06} & \metricstdcorner{0.8962}{0.0009} & \metricstdcorner{0.0773}{0.0018} & 166 & \metricstdcorner{24.08}{0.07} & \metricstdcorner{0.8532}{0.0012} & \metricstdcorner{0.2398}{0.0009} & 147 \\
    ESC-DGR & \metricstdcorner{\textbf{28.96}}{0.04} & \metricstdcorner{\textbf{0.9229}}{0.0011} & \metricstdcorner{\textbf{0.0612}}{0.0021} & \underline{178} & \metricstdcorner{\textbf{31.62}}{0.11} & \metricstdcorner{\textbf{0.9237}}{0.0011} & \metricstdcorner{\textbf{0.0712}}{0.0012} & \underline{183} \\
    \bottomrule
  \end{tabularx}
\end{table*}

\begin{table}[width=\columnwidth,pos=t]
  \centering
  \caption{Quantitative comparisons in dynamic regions of synthetic and real scenes. Results are grouped by scene domain. Mean denotes the arithmetic average over scenes within each group. The small $\pm$ values shown at the lower right of each mean denote standard deviations. Leading zeros are omitted for values below 1. Bold indicates the best value in each row.}
  \label{tab:dynamic-region}
  \rmfamily
  \fontsize{7.2}{8.7}\selectfont
  \setlength{\tabcolsep}{1pt}
  \setlength{\heavyrulewidth}{.5pt}
  \setlength{\lightrulewidth}{.3pt}
  \setlength{\cmidrulewidth}{.3pt}
  \setlength{\arrayrulewidth}{.3pt}
  \renewcommand{\arraystretch}{1.22}
  \setlength{\aboverulesep}{2pt}
  \setlength{\belowrulesep}{2pt}
  \begin{tabularx}{\columnwidth}{@{}>{\centering\arraybackslash}m{.15\columnwidth}|YYY|YYY@{}}
    \toprule
    \multirow{2}{*}{Scene} & \multicolumn{3}{c|}{E-D3DGS~\cite{ref14}} & \multicolumn{3}{c}{ESC-DGR} \\
    \cmidrule(lr){2-4}\cmidrule(lr){5-7}
    & dPSNR$\uparrow$ & dSSIM$\uparrow$ & dLPIPS$\downarrow$ & dPSNR$\uparrow$ & dSSIM$\uparrow$ & dLPIPS$\downarrow$ \\
    \midrule
    \multicolumn{7}{@{}l}{\textit{Synthetic scenes}} \\
    \midrule
    Lego & \metricstdcorner{23.65}{.10} & \metricstdcorner{.9485}{.0012} & \metricstdcorner{.0252}{.0010} & \metricstdcorner{\textbf{24.83}}{.12} & \metricstdcorner{\textbf{.9756}}{.0020} & \metricstdcorner{\textbf{.0080}}{.0001} \\
    Hotdog & \metricstdcorner{26.72}{.05} & \metricstdcorner{.9303}{.0008} & \metricstdcorner{.0238}{.0004} & \metricstdcorner{\textbf{27.47}}{.07} & \metricstdcorner{\textbf{.9609}}{.0009} & \metricstdcorner{\textbf{.0117}}{.0001} \\
    Materials & \metricstdcorner{30.44}{.07} & \metricstdcorner{.9917}{.0009} & \metricstdcorner{.0035}{.0001} & \metricstdcorner{\textbf{31.21}}{.08} & \metricstdcorner{\textbf{.9985}}{.0009} & \metricstdcorner{\textbf{.0018}}{.0001} \\
    Musicbox & \metricstdcorner{23.95}{.07} & \metricstdcorner{.9552}{.0008} & \metricstdcorner{.0266}{.0005} & \metricstdcorner{\textbf{24.91}}{.11} & \metricstdcorner{\textbf{.9632}}{.0010} & \metricstdcorner{\textbf{.0210}}{.0003} \\
    Celestial globe & \metricstdcorner{19.54}{.04} & \metricstdcorner{.9785}{.0006} & \metricstdcorner{.0151}{.0002} & \metricstdcorner{\textbf{20.11}}{.05} & \metricstdcorner{\textbf{.9843}}{.0010} & \metricstdcorner{\textbf{.0123}}{.0001} \\
    Fan & \metricstdcorner{21.57}{.04} & \metricstdcorner{.9501}{.0012} & \metricstdcorner{.0255}{.0002} & \metricstdcorner{\textbf{22.51}}{.06} & \metricstdcorner{\textbf{.9589}}{.0002} & \metricstdcorner{\textbf{.0224}}{.0002} \\
    Waterwheel & \metricstdcorner{20.74}{.05} & \metricstdcorner{.9690}{.0008} & \metricstdcorner{.0168}{.0003} & \metricstdcorner{\textbf{22.93}}{.08} & \metricstdcorner{\textbf{.9762}}{.0011} & \metricstdcorner{\textbf{.0132}}{.0001} \\
    Man & \metricstdcorner{21.03}{.06} & \metricstdcorner{.9492}{.0008} & \metricstdcorner{.0241}{.0003} & \metricstdcorner{\textbf{21.89}}{.08} & \metricstdcorner{\textbf{.9601}}{.0015} & \metricstdcorner{\textbf{.0209}}{.0002} \\
    \midrule
    Mean & \metricstdcorner{23.46}{.03} & \metricstdcorner{.9591}{.0006} & \metricstdcorner{.0201}{.0002} & \metricstdcorner{\textbf{24.48}}{.03} & \metricstdcorner{\textbf{.9722}}{.0004} & \metricstdcorner{\textbf{.0139}}{.0001} \\
    \midrule
    \multicolumn{7}{@{}l}{\textit{Real-world scenes}} \\
    \midrule
    Excavator & \metricstdcorner{21.61}{.05} & \metricstdcorner{.9583}{.0009} & \metricstdcorner{.0188}{.0003} & \metricstdcorner{\textbf{23.35}}{.08} & \metricstdcorner{\textbf{.9668}}{.0011} & \metricstdcorner{\textbf{.0174}}{.0002} \\
    Jeep & \metricstdcorner{20.81}{.06} & \metricstdcorner{.9281}{.0007} & \metricstdcorner{.0282}{.0005} & \metricstdcorner{\textbf{22.47}}{.10} & \metricstdcorner{\textbf{.9754}}{.0010} & \metricstdcorner{\textbf{.0262}}{.0003} \\
    Flowers & \metricstdcorner{20.75}{.07} & \metricstdcorner{.9738}{.0008} & \metricstdcorner{.0177}{.0003} & \metricstdcorner{\textbf{21.83}}{.07} & \metricstdcorner{\textbf{.9865}}{.0013} & \metricstdcorner{\textbf{.0151}}{.0001} \\
    Bird & \metricstdcorner{22.42}{.04} & \metricstdcorner{.9682}{.0007} & \metricstdcorner{.0253}{.0003} & \metricstdcorner{\textbf{23.74}}{.06} & \metricstdcorner{\textbf{.9797}}{.0008} & \metricstdcorner{\textbf{.0196}}{.0002} \\
    \midrule
    Mean & \metricstdcorner{21.40}{.03} & \metricstdcorner{.9571}{.0004} & \metricstdcorner{.0225}{.0002} & \metricstdcorner{\textbf{22.85}}{.04} & \metricstdcorner{\textbf{.9771}}{.0004} & \metricstdcorner{\textbf{.0196}}{.0001} \\
    \bottomrule
  \end{tabularx}
\end{table}

\subsection{Comparison with existing methods}

\begin{figure*}[width=0.92\textwidth,pos=!t]
  \centering
  \includegraphics[width=\textwidth,trim=0 32bp 0 0,clip]{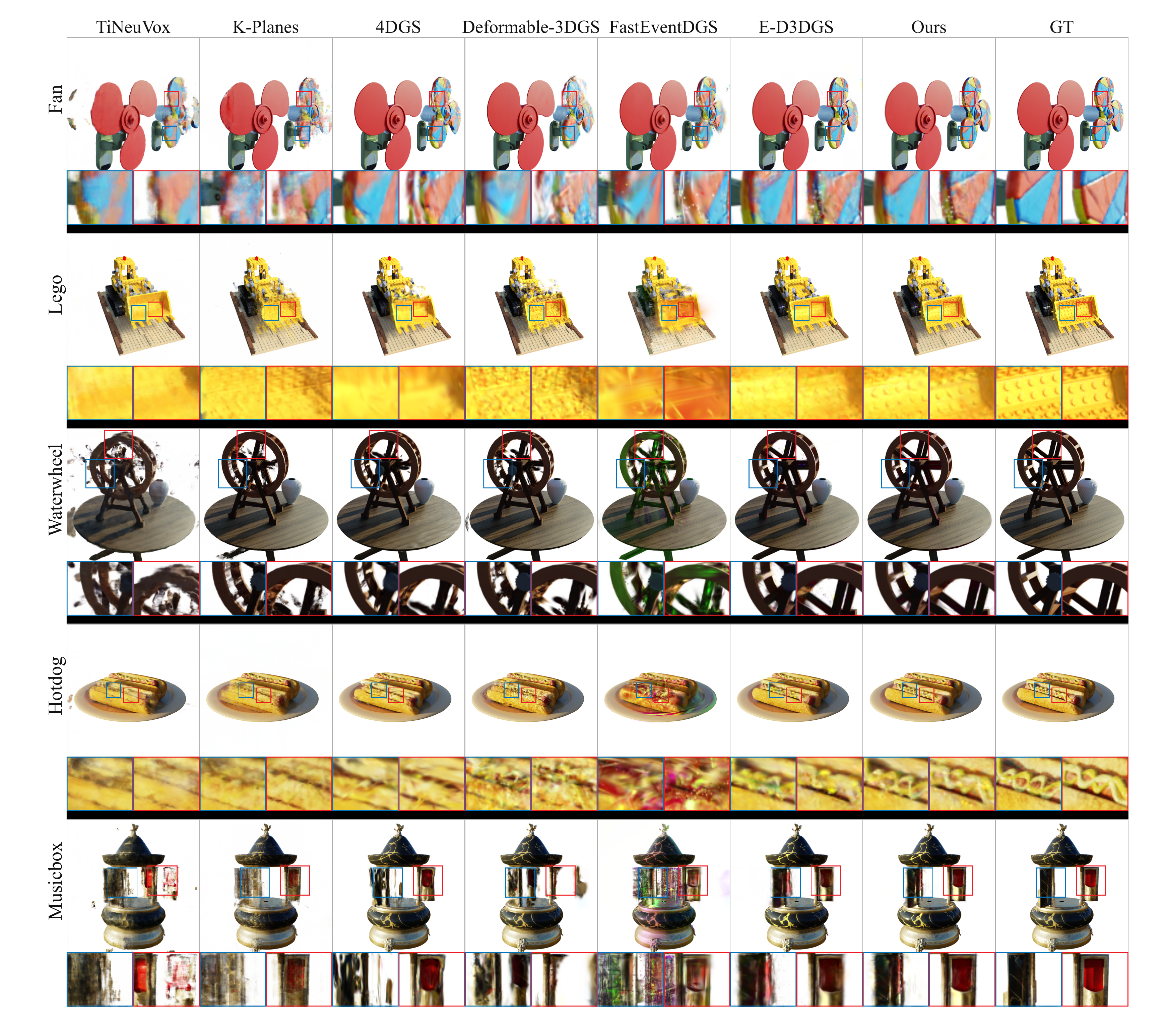}
  \caption{Qualitative comparisons on the synthetic dataset. The enlarged views below correspond to the blue and red boxes in the main images, respectively, showing motion boundaries and local texture details.}
  \label{fig:5}
\end{figure*}

Fig. \ref{fig:5} compares full images and local regions at the same viewpoints and times. Tables \ref{tab:2} and \ref{tab:real-world} report full-image metrics for synthetic and real scenes, respectively. Table \ref{tab:dynamic-region} summarizes dynamic-region metrics for both domains. We analyze results in terms of boundary and texture preservation, structural completeness, and local artifacts.

\subsubsection{Qualitative comparisons}
\label{sec:qual}

\begin{figure*}[width=0.92\textwidth,pos=t]
  \centering
  \makebox[\textwidth][l]{\hspace*{0.028918919\textwidth}\includegraphics[width=0.938588589\textwidth]{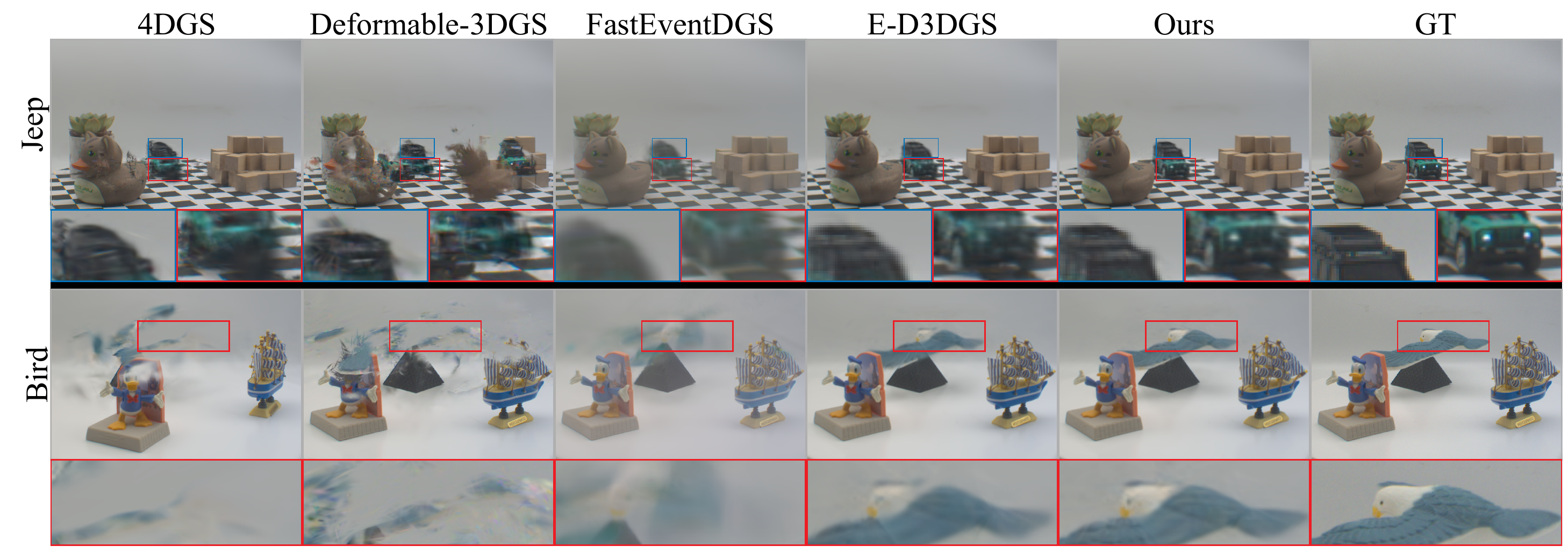}}
  \caption{Qualitative comparisons on real scenes. The left and right enlarged views below Jeep correspond to the blue and red boxes in the main image, respectively. The enlarged view below Bird corresponds to the red box. These views show local vehicle structures and bird contour details.}
  \label{fig:real-qualitative}
\end{figure*}

Fast-moving regions in Fan, Hotdog, and Lego show color bleeding, excessive smoothing, and grainy artifacts (Fig. \ref{fig:5}). Our reconstructions preserve sharper color boundaries on the fan blades, a more continuous yellow curve, and more regular protrusions on the bucket. Similar improvements appear in Jeep and Bird (Fig. \ref{fig:real-qualitative}). Vehicle edges are more clearly separated from adjacent components in the enlarged Jeep views. Color bleeding between the bird silhouette and nearby background is also reduced. Both groups exhibit clearer motion boundaries and local detail, although some grainy artifacts remain near the Fan blades and Hotdog textures.

In Waterwheel, our method separates the reconstructed rim, spokes, and connecting components more clearly, with fewer color artifacts inside the wheel and fewer spurious responses in the gaps. Vertical structures and foreground boundaries in Musicbox are also more stable. The vehicle body and local components in Jeep retain more coherent shapes, and reduced smearing and ghosting around the moving bird yield a cleaner silhouette in Bird. The improvements primarily concern motion-boundary continuity, component separation, and local artifact suppression; some blur remains around the Waterwheel hub and smaller components.

FastEventDGS produces pronounced color bleeding and ghosting around the Lego bucket, smearing in Hotdog textures, color artifacts near the Waterwheel spokes, and blurred vertical structures in Musicbox (Fig. \ref{fig:5}). Local vehicle components in Jeep and the bird silhouette in Bird are also blurred (Fig. \ref{fig:real-qualitative}). Our reconstructions more clearly distinguish local components, motion boundaries, and textures. These qualitative comparisons use the same RGB-supervised adaptation of FastEventDGS as the full-image quantitative evaluation.

These results show the joint contribution of support augmentation, structural guidance, and local refinement to dynamic contour and texture recovery.

\subsubsection{Quantitative comparisons}
\label{sec:quant}

Tables \ref{tab:2} and \ref{tab:real-world} report full-image metrics for synthetic and real scenes, respectively. Table \ref{tab:dynamic-region} groups dynamic-region metrics in the same way. On the eight synthetic scenes, the five RGB-only baselines are constrained by sparse training observations. Their 8-scene mean full-image PSNR is 5.8--10.0~dB below that of E-D3DGS (Table~\ref{tab:2}). We therefore report gains over E-D3DGS when comparing event-based methods.

Our method outperforms FastEventDGS in full-image PSNR, SSIM, and LPIPS across all twelve scenes. FastEventDGS achieves average full-image PSNR values of 24.41 and 26.08~dB on synthetic and real scenes, respectively, with average frame rates of 132 and 155~FPS. This comparison uses the RGB-supervised adaptation described in Section \ref{sec:experiment-settings}.

Relative to E-D3DGS, our method improves both full-image and dynamic-region reconstruction metrics in every scene. On synthetic scenes, mean full-image PSNR increases by 0.42~dB. Mean dPSNR rises from 23.46 to 24.48~dB, a difference of 1.02~dB between the displayed group means; on real scenes, the corresponding difference is 1.45~dB. Both dSSIM and dLPIPS also improve. These results are consistent with the complementary roles of the three modules: EDSA supplies missing support, CCSMG provides structural references for weakly observed Gaussians, and EHMR corrects rapid inter-frame displacements. Gains vary with motion magnitude and structural scale, with the clearest improvements in Man, Waterwheel, and Excavator.

Our method maintains rendering speeds above 80~FPS in all twelve scenes, preserving real-time performance at a lower speed than the direct baseline. The mean frame rate decreases from 202 to 159~FPS on synthetic scenes, a reduction of about 21\%, and from 209 to 158~FPS on real scenes, a reduction of about 25\%. Mean frame rates nevertheless remain higher than those of 4DGS and Deformable-3DGS.

\subsubsection{Gaussian count and representation compactness}

Across the eight synthetic scenes, our method uses fewer dynamic Gaussians in seven of the eight scenes, reducing the total by 8.5\%, from 378,030 to 345,887. Lego and Materials each show reductions of more than one fifth. Despite these lower counts, both full-image and dynamic-region reconstruction metrics remain better than E-D3DGS.

In Lego, support injection adds 1,762 dynamic Gaussians at iteration 3,001. Subsequent net growth to iteration 18,000 is 17,485, compared with 24,636 for E-D3DGS. This difference of 7,151 Gaussians is approximately four times the number injected. It leaves 5,389 fewer Gaussians when densification ends. Thus, the final reduction mainly reflects subsequent densification and pruning. The geometry, appearance, and opacity of the added Gaussians remain trainable. These Gaussians are removed if they do not meet the pruning criteria. A possible explanation is that augmented support distributes errors more broadly, reducing repeated gradient-triggered splitting.

\begin{figure}[width=\columnwidth,pos=htbp]
  \centering
  \includegraphics[width=\columnwidth,trim=0 4.5bp 0 0,clip]{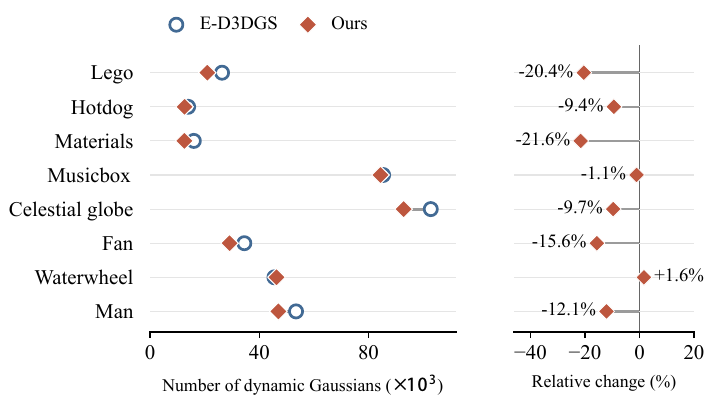}
  \caption{Dynamic Gaussian counts across eight synthetic scenes. Right: relative change in count compared with E-D3DGS, $100(N_{\mathrm{ours}}-N_{\mathrm{base}})/N_{\mathrm{base}}$.}
  \label{fig:gaussian-count}
\end{figure}

\begin{figure}[width=\columnwidth,pos=htbp]
  \centering
  \includegraphics[width=0.96\columnwidth]{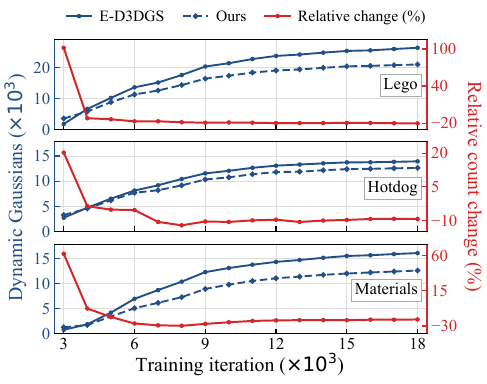}
  \caption{Evolution of dynamic Gaussian counts during training. The right axis shows relative count change, $100(N_{\mathrm{ours}}-N_{\mathrm{base}})/N_{\mathrm{base}}$ (\%).}
  \label{fig:gaussian-trajectories}
\end{figure}

Fig. \ref{fig:gaussian-trajectories} shows that the relative count falls rapidly after iteration 3,001, drops below the baseline by iteration 4,000, and then decreases more gradually. Hotdog and Materials show similar trends of different magnitudes. These observations are consistent with reduced densification pressure after support injection, rather than a prescribed compression ratio. Count reduction is not itself an objective: Waterwheel uses 1.6\% more dynamic Gaussians and gains 2.19~dB in dPSNR, reflecting the support needed for its thin, rapidly rotating spokes.

\begin{table*}[width=\textwidth,pos=t]
  \centering
  \caption{Ablations with cumulative module additions, module combinations, and full structural graph replacement across twelve scenes (eight synthetic and four real). Each metric is the arithmetic mean over the twelve scenes. The small $\pm$ values shown at the lower right of each mean denote standard deviations. Leading zeros are retained for means below 1. The best means are marked in bold.}
  \label{tab:main-ablation}
  \rmfamily
  \fontsize{8}{9.5}\selectfont
  \setlength{\tabcolsep}{2pt}
  \renewcommand{\arraystretch}{1.22}
  \begin{tabularx}{\textwidth}{@{}>{\centering\arraybackslash}m{\dimexpr.27\textwidth-6pt\relax}ccc@{\hspace{8pt}}|YYYY|YYY}
    \toprule
    \multirow[c]{2}{*}[-0.55ex]{Method} & \multirow[c]{2}{*}[-0.55ex]{$P$} & \multirow[c]{2}{*}[-0.55ex]{$G$} & \multirow[c]{2}{*}[-0.55ex]{$H$} & \multicolumn{4}{c|}{Full image} & \multicolumn{3}{c}{Dynamic region} \\
    \cmidrule(l{4pt}r{4pt}){5-8}\cmidrule(l{4pt}r{4pt}){9-11}
    & & & & PSNR$\uparrow$ & SSIM$\uparrow$ & LPIPS$\downarrow$ & FPS$\uparrow$ & dPSNR$\uparrow$ & dSSIM$\uparrow$ & dLPIPS$\downarrow$ \\
    \midrule
    E-D3DGS & -- & -- & -- & \metricstdcorner{31.13}{0.05} & \metricstdcorner{0.9474}{0.0006} & \metricstdcorner{0.0403}{0.0004} & 204 & \metricstdcorner{22.77}{0.03} & \metricstdcorner{0.9584}{0.0006} & \metricstdcorner{0.0209}{0.0002} \\
    with EDSA only & $\checkmark$ & -- & -- & \metricstdcorner{31.26}{0.04} & \metricstdcorner{0.9466}{0.0005} & \metricstdcorner{0.0400}{0.0003} & \textbf{208} & \metricstdcorner{23.27}{0.03} & \metricstdcorner{0.9606}{0.0005} & \metricstdcorner{0.0212}{0.0002} \\
    w/o EDSA & -- & $\checkmark$ & $\checkmark$ & \metricstdcorner{31.38}{0.02} & \metricstdcorner{0.9505}{0.0004} & \metricstdcorner{0.0371}{0.0003} & 154 & \metricstdcorner{23.32}{0.03} & \metricstdcorner{0.9689}{0.0006} & \metricstdcorner{0.0175}{0.0003} \\
    w/o CCSMG & $\checkmark$ & -- & $\checkmark$ & \metricstdcorner{31.33}{0.05} & \metricstdcorner{0.9479}{0.0005} & \metricstdcorner{0.0388}{0.0002} & 176 & \metricstdcorner{23.49}{0.05} & \metricstdcorner{0.9675}{0.0005} & \metricstdcorner{0.0186}{0.0002} \\
    Adapted USplat4D graph & $\checkmark$ & $\ast$ & $\checkmark$ & \metricstdcorner{31.41}{0.02} & \metricstdcorner{0.9507}{0.0006} & \metricstdcorner{0.0366}{0.0003} & 164 & \metricstdcorner{23.68}{0.02} & \metricstdcorner{0.9693}{0.0006} & \metricstdcorner{0.0179}{0.0004} \\
    w/o EHMR & $\checkmark$ & $\checkmark$ & -- & \metricstdcorner{31.44}{0.04} & \metricstdcorner{0.9484}{0.0005} & \metricstdcorner{0.0379}{0.0002} & 190 & \metricstdcorner{23.60}{0.05} & \metricstdcorner{0.9642}{0.0005} & \metricstdcorner{0.0203}{0.0003} \\
    ESC-DGR & $\checkmark$ & $\checkmark$ & $\checkmark$ & \metricstdcorner{\textbf{31.58}}{0.02} & \metricstdcorner{\textbf{0.9546}}{0.0004} & \metricstdcorner{\textbf{0.0343}}{0.0002} & 159 & \metricstdcorner{\textbf{23.94}}{0.03} & \metricstdcorner{\textbf{0.9738}}{0.0004} & \metricstdcorner{\textbf{0.0158}}{0.0001} \\
    \bottomrule
  \end{tabularx}
  \par\vspace{3pt}
  {\footnotesize\rmfamily\raggedright Note: $P$, $G$, and $H$ denote EDSA, CCSMG, and EHMR, respectively. A $\checkmark$ indicates an enabled module, and -- indicates a disabled module. An $\ast$ indicates replacement of CCSMG with an adapted USplat4D structural graph. The ``with EDSA only'' variant enables only EDSA on the E-D3DGS baseline. This row, ``w/o EHMR,'' and ESC-DGR form a cumulative comparison that progressively adds modules. The adapted graph includes analytical RGB uncertainty, anisotropic 3D uncertainty, UA-KNN, DQB motion references, and uncertainty-weighted Key/Non-key objectives. All other training and evaluation settings are kept unchanged.\par}
\end{table*}

\subsection{Ablation studies}

Table \ref{tab:main-ablation} summarizes cumulative module additions, module combinations, and replacement of the full structural graph across twelve scenes (eight synthetic and four real). E-D3DGS serves as the baseline without the three proposed modules. The ``with EDSA only,'' ``w/o EHMR,'' and ESC-DGR variants successively add EDSA, CCSMG, and EHMR, while three module-removal variants assess each component within the full framework. Along the cumulative sequence, mean dPSNR rises from 22.77 to 23.27, 23.60, and 23.94~dB, and full-image PSNR rises from 31.13 to 31.26, 31.44, and 31.58~dB. We first examine individual modules and then their interactions.

\begin{figure}[width=\columnwidth,pos=htbp]
  \centering
  \includegraphics[width=\columnwidth,trim=0 44bp 0 0,clip]{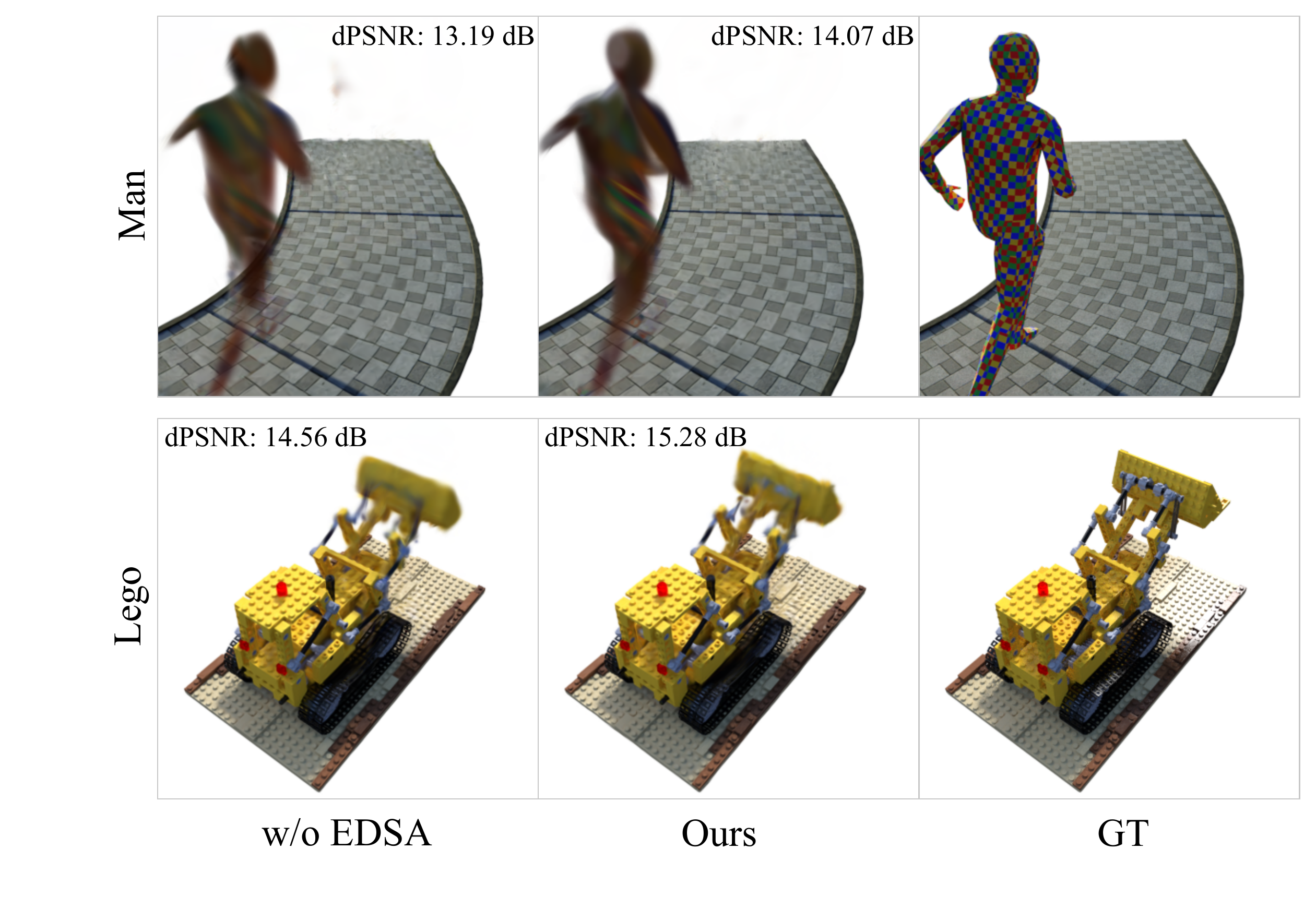}
  \caption{Early-training ablation of Event-guided Dynamic Support Augmentation (EDSA). At iteration 4,000, dynamic-region dPSNR increases by 0.88 and 0.72~dB in Man and Lego, respectively. Removing EDSA produces missing structures and ghosting at dynamic boundaries.}
  \label{fig:edsa-early-ablation}
\end{figure}

Removing EDSA reduces twelve-scene mean dPSNR by 0.62~dB, the largest decrease among the three module removals. At iteration 4,000, EDSA already improves Man and Lego dPSNR by 0.88 and 0.72~dB, respectively, with more complete dynamic boundaries (Fig. \ref{fig:edsa-early-ablation}). Neither graph guidance nor residual refinement is active then, so these early gains are attributable to support augmentation. Count differences before densification ends at iteration 18,000 likewise arise only from this strategy: the initial injection is followed by lower net growth, ultimately reducing total dynamic count by 8.5\% across eight synthetic scenes (Fig. \ref{fig:gaussian-count}), with reductions in seven.

EDSA alone improves dPSNR by 0.50~dB, dSSIM by 0.0022, and full-image PSNR by 0.13~dB, with rendering speed largely unchanged. Changes in full-image SSIM and dLPIPS are comparable to seed variation. The quality gain therefore comes with fewer primitives, not more. Together with the removal result, this supports both its standalone contribution and its role in the full model.

\begin{figure}[width=\columnwidth,pos=htbp]
  \centering
  \includegraphics[width=\columnwidth,trim=0 5bp 0 0,clip]{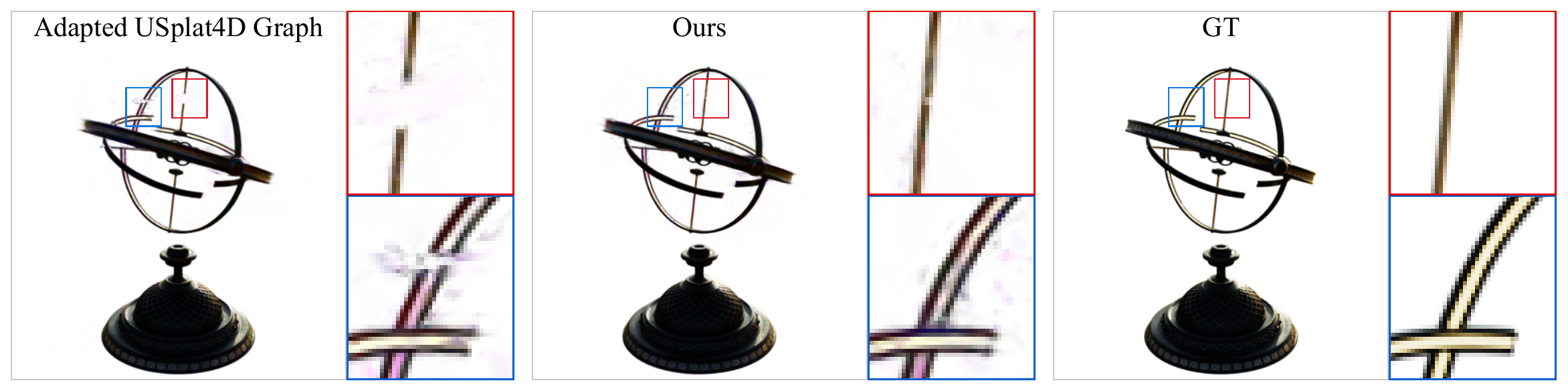}
  \caption{Qualitative comparison of structural graph designs in Celestial globe. Red and blue boxes show enlarged views of vertical and curved rods, respectively. The adapted USplat4D graph produces broken rods and spurious responses. Our method preserves more continuous rods with fewer artifacts.}
  \label{fig:usplat-graph-ablation}
\end{figure}

Adding CCSMG to EDSA improves full-image PSNR by 0.18~dB and dPSNR by 0.33~dB; dSSIM rises from 0.9606 to 0.9642 and dLPIPS falls from 0.0212 to 0.0203. Removing it from the full model reduces full-image PSNR by 0.25~dB, the largest decrease among module removals. Replacing CCSMG with the adapted USplat4D graph lowers dPSNR from 23.94 to 23.68~dB and produces broken or spurious rod structures (Fig. \ref{fig:usplat-graph-ablation}). This comparison retains EDSA, EHMR, and single-stage training, suggesting that reliable reference selection and propagation matter beyond the Key/Non-key framework itself.

Adding EHMR after EDSA and CCSMG improves dPSNR by 0.34~dB, raises dSSIM from 0.9642 to 0.9738, and reduces dLPIPS from 0.0203 to 0.0158 (22.2\%). The structural and perceptual gains exceed those of either preceding cumulative step. Removing EHMR likewise causes the largest losses in these two metrics. Event-conditioned residuals therefore particularly benefit structural similarity and perceptual quality in dynamic regions. These gains come at the cost of speed: the frame rate drops from 190 to 159~FPS.

EDSA improves mean dPSNR by 0.50~dB when used alone and by 0.62~dB when added to CCSMG and EHMR. The full model gains 1.17~dB over the baseline, compared with the summed gains of 1.05~dB from EDSA alone and CCSMG--EHMR without EDSA. The resulting 0.12~dB difference in the reported means is consistent with complementary effects between support augmentation and motion modeling. This is a descriptive contrast between reported means. It does not establish statistical significance or stability across random seeds. The larger benefit of EDSA within the full model suggests that additional support can be used more effectively when accompanied by structural guidance and event-conditioned refinement.

With EDSA enabled, adding CCSMG or EHMR alone improves dPSNR by 0.33 or 0.22~dB, respectively, whereas adding both yields a 0.67~dB gain, slightly exceeding their summed individual gains of 0.55~dB. Equivalently, the gain from EHMR increases from 0.22 to 0.34~dB in the presence of CCSMG, while the gain from CCSMG increases from 0.33 to 0.45~dB in the presence of EHMR. A similar pattern appears in dynamic-region structural and perceptual metrics: the joint dSSIM improvement exceeds the summed individual improvements by 0.0027, and the joint dLPIPS reduction exceeds them by 0.0019. These differences in the reported means are consistent with structural guidance and event-conditioned refinement benefiting one another. Establishing the stability of this interaction would require computing the contrast separately for each of the shared random seeds and assessing its variation.

The ablations test the dependencies motivating the method. EDSA improves reconstruction without either motion component, showing that event-guided placement of support has independent value. Its larger gain within the full model is consistent with the benefit of learning motion on that support. Likewise, the increased benefit of EHMR in the presence of CCSMG is consistent with refining a structurally guided estimate rather than recovering the full motion from sparse events. These observations are consistent with the proposed dependencies, although the stability of the interaction effects across random seeds remains to be established.

\subsection{Limitations}
\label{sec:limits}

ESC-DGR improves reconstruction quality at the cost of rendering speed and dependence on synchronized events. Mean frame rates decrease by approximately 21\% and 25\% on synthetic and real scenes, respectively, although rendering remains above 80~FPS in all twelve scenes. Full-model inference requires temporally aligned target-time events. When these events are unavailable or poorly synchronized, EHMR must be disabled. This variant achieves a mean dPSNR of 23.60~dB across twelve scenes, compared with 22.77~dB for E-D3DGS and 23.94~dB for the full model, providing a trade-off between speed and reconstruction quality.

ESC-DGR provides smaller gains when errors are dominated by appearance and texture rather than insufficient support or rapid inter-frame motion. The dPSNR gains in Celestial globe, Hotdog, and Materials range from 0.57 to 0.77~dB, compared with 2.19~dB in Waterwheel. Grainy artifacts remain near the Fan blades and Hotdog textures, and some blur persists around the Waterwheel hub and smaller components. Sparse events in weakly textured or low-contrast regions may exclude moving Gaussians from the active set. The residual bound in Eq. \eqref{eq:23} also limits the correction of misalignments beyond the local Gaussian scale. These limitations constrain the benefits of local refinement even when the overall reconstruction improves.

\section{Conclusion}

Fast motion invalidates two assumptions in dynamic Gaussian reconstruction. The first is that existing Gaussians can absorb reconstruction errors. The second is that photometric optimization has converged sufficiently to estimate reliability using analytical variance. ESC-DGR uses event observations independent of these conditions to locate additional support and complement online rendering statistics in selecting motion references. Support allows the representation to explain the observations, while reliable references guide motion in weakly observed regions. Target-time events then refine rapid local displacements that the frame-level estimate does not capture. This connects support placement and motion refinement within one evolving Gaussian representation.

For event-driven monocular dynamic scene reconstruction, ESC-DGR achieves state-of-the-art reconstruction quality on the twelve-scene E-D3DGS benchmark. Relative to E-D3DGS, synthetic and real dPSNR improve by 1.02 and 1.45~dB, respectively, and the total dynamic Gaussian count across eight synthetic scenes decreases by 8.5\%. The gains come at lower rendering speeds: average frame rates decrease by approximately 21\% and 25\%, respectively, while remaining above 80~FPS in all scenes. Full-model inference also requires synchronized target-time events. These costs define the trade-off accompanying the reconstruction improvements.

\section*{Declarations}

Funding: This work was supported by the National Natural Science Foundation of China under Grant No.~62473210 and the Natural Science Foundation of Tianjin under Grant Nos.~24JCYBJC00240 and 2025JCQNJC01350.

\subsection*{Declaration of competing interest}

The authors declare that they have no known competing financial interests or personal relationships that could have appeared to influence the work reported in this paper.

Data availability: The research data and evaluation materials supporting this study are available on Zenodo at \url{https://doi.org/10.5281/zenodo.24296657}.

Code availability: The source code and training configurations for this study are publicly available at \url{https://github.com/zh-yu1/ESC-DGR}.

\subsection*{CRediT authorship contribution statement}

\textbf{Zihao Yu:} Writing -- original draft, Methodology, Software, Investigation, Validation, Visualization.
\textbf{Guozhen Liu:} Methodology, Software, Investigation, Validation, Writing -- review \& editing.
\textbf{Jie Yan:} Validation, Formal analysis, Writing -- review \& editing.
\textbf{Jialiang Liu:} Data curation, Validation, Visualization.
\textbf{Panlong Tan:} Funding acquisition, Resources.
\textbf{Qinglin Sun:} Supervision, Funding acquisition.
\textbf{Hao Sun:} Conceptualization, Supervision, Writing -- review \& editing.

\subsection*{Declaration of generative AI and AI-assisted technologies in the manuscript preparation process}

During the preparation of this work, the authors used OpenAI Codex to polish the language and improve the clarity and readability of the manuscript. After using this tool, the authors reviewed and edited the content as needed and take full responsibility for the content of the published article.

\FloatBarrier
\bibliographystyle{elsarticle-num}
\bibliography{references}
\balance

\clearpage
\onecolumn
\appendix
\numberwithin{equation}{section}
\counterwithin{table}{section}
\counterwithin{figure}{section}
\section{Notation and parameter settings}
\label{app:parameters}

This appendix summarizes the main notation and shared parameters. For scene-dependent parameters, we give ranges or relationships where specified and otherwise mark them as scene-specific; this table does not enumerate the configuration of each scene. Lengths are expressed in canonical scene coordinates, and $k$ denotes the main training iteration, excluding threshold-network pretraining.

\begingroup
\small
\setlength{\tabcolsep}{5pt}
\renewcommand{\arraystretch}{1.12}
\captionof{table}{Main notation, shared parameters, and adjustment rules.}
\label{tab:parameter-common}
\noindent\begin{tabularx}{\textwidth}{@{}p{0.25\textwidth}X p{0.25\textwidth}c@{}}
\toprule
Symbol & Description & Value & Scene-dependent\\
\midrule
$\lambda_g,\lambda_c$ & Mixing weights for appearance boundaries and temporal reliability & $0.5,\ 0.5$ & No\\
$h_v,N_{\mathrm{sup}}$ & Event aggregation voxel size and maximum candidate count & $0.01$; $N_{\mathrm{sup}}$ scene-specific & Partly\\
$\tau_{\psi},n_{\min}$ & Voxel score threshold and minimum candidate count & $0.15$--$0.35,\ 1$--$3$ & Yes\\
$\lambda_{\mathrm{ema}},n_0,b_{\mathrm{err}}$ & Contribution EMA coefficient, observation saturation count, and error decay scale & $0.9,\ 5,\ 0.12$ & No\\
$(\beta_U,\beta_V,\beta_{\mathrm{rgb}},\beta_{\mathrm{evt}})$ & Fusion weights for contribution, visibility, RGB quality, and event support & $(0.45,0.25,0.20,0.10)$ & No\\
$\tau_C,\tau_Q,o_{\min}$ & Key contribution threshold, reliability threshold, and minimum observation count & $0.08,\ 0.35$--$0.50,\ 5$--$10$ & Partly\\
$r_{\mathrm{key}},N_{\mathrm{key}}^{\min},N_{\mathrm{key}}^{\max}$ & Target Key ratio and minimum and maximum Key counts & $0.04,\ 256$; scene-specific maximum & Partly\\
$n_{\mathrm{nb}},h_{\mathrm{graph}},d_{\max}$ & Graph neighbor count, spatial decay scale, and maximum valid edge distance & $n_{\mathrm{nb}},h_{\mathrm{graph}}$ scene-specific; $d_{\max}=3h_{\mathrm{graph}}$ & Yes\\
$\varepsilon_C,\varepsilon_Q,\gamma_r$ & Lower bounds on propagation weights and Non-key residual scale & $0.10,\ 0.25,\ 0.25$ & No\\
$\lambda_a,\lambda_q,\lambda_t,\lambda_{res}$ & Graph weights for translation anchoring, rotation anchoring, temporal regularization, and residuals & $0.1,\ 0.1,\ 0.05,\ 10^{-4}$ & No\\
$\lambda_n,\lambda_r,\lambda_G^{\max}$ & Weights for Non-key consistency, rigidity, and the total graph loss &  scene-specific & Yes\\
$\gamma_s,\ell_{\min},\ell_{\max}$ & Scale factor and lower and upper bounds for the high-frequency residual radius & $0.75,\ 0.001,\ 0.05$ & No\\
$s_{\mathrm{hi}}^{\max},\lambda_{\mathrm{th}}$ & Final high-frequency residual scale for RGB rendering and threshold calibration weight & $0.005,\ 5$ & No\\
$\lambda_{\mathrm{evt}}^{\max},\lambda_{mag}^{\max}$ & Weights for event supervision and residual magnitude regularization & $5+0.05,\ 10^{-5}$ & No\\
$\lambda_d,\lambda_b,\lambda_{art}$ & Auxiliary weights for dynamic RGB, edges, and opacity & Scene-specific; set to 0 to disable & Yes\\
\bottomrule
\end{tabularx}
\endgroup

Graph constraints, high-frequency residuals, and auxiliary losses are introduced progressively according to the training stages described in the main text, with their strengths adjusted during training. The ranking constants 0.25 and 0.02 and the observation-count cap of 5 are fixed empirical settings that limit the influence of reliability and observation count relative to the primary contribution term; they are not adjusted per scene.

\clearpage
\section{Implementation and evaluation details}
\label{app:implementation}

\subsection{Gaussian kernels and covariance}
\label{app:gaussian}

Here, $\mathcal G$ is the Gaussian set, $\boldsymbol\mu_i\in\mathbb R^3$ is the center, and $\mathbf R_i\in\mathbb R^{3\times3}$ is the rotation matrix. The diagonal matrix $\mathbf S_i=\operatorname{diag}(s_{i,1},s_{i,2},s_{i,3})$ contains the three axial scales, with $s_{i,d}>0$ denoting the scale along axis $d$. The opacity is $\alpha_i\in[0,1]$, and the RGB color $\mathbf c_i$ is determined by spherical harmonic color parameters and the viewing direction. The spatial kernel of Gaussian $i$ at $\mathbf x\in\mathbb R^3$ is defined as

\begin{equation}
\varphi_i(\mathbf x)=
\exp\!\left[-\frac{1}{2}
(\mathbf x-\boldsymbol\mu_i)^{\mathsf T}
\boldsymbol\Sigma_i^{-1}
(\mathbf x-\boldsymbol\mu_i)\right].
\label{eq:gaussian-kernel}
\end{equation}

Here, $\varphi_i$ is the unnormalized Gaussian spatial response, which decays as $\mathbf x$ moves away from the center. The positive-definite covariance matrix $\boldsymbol\Sigma_i\in\mathbb R^{3\times3}$ determines the shape and orientation of this response in 3D space. To ensure a valid covariance matrix and facilitate optimization, we parameterize it as

\begin{equation}
\boldsymbol\Sigma_i=\mathbf R_i\mathbf S_i\mathbf S_i^{\mathsf T}\mathbf R_i^{\mathsf T}.
\label{eq:2}
\end{equation}

The superscript $\mathsf T$ denotes matrix transpose. During rasterization, the camera model projects each 3D kernel onto a 2D elliptical kernel, which combines with the Gaussian opacity to define the effective alpha at each pixel.

A deformation field maps canonical Gaussians to time $t$\cite{ref6,ref7}. In our implementation, the low-frequency branch predicts translation, rotation, and scale offsets, while the high-frequency branch adds position residuals. The projected Gaussians are sorted by depth and alpha-composited according to Eq. \eqref{eq:3}.

\subsection{Event normalization and image edges}
\label{app:normalization}

Let $\mathcal D_f^{\mathrm{pix}}=\{0,\ldots,W_f-1\}\times\{0,\ldots,H_f-1\}$ denote the full pixel domain of frame $f$. To measure the difference between current event activity and its temporally fused result, we define the temporal difference map $D_f^{\mathrm{temp}}$ and its within-frame normalized version $\widetilde D_f^{\mathrm{temp}}$:
\begin{equation}
\begin{aligned}
D_f^{\mathrm{temp}}(\mathbf p)
&=\bigl|\bar A_f(\mathbf p)-A_f(\mathbf p)\bigr|,\\
D_{f,-}^{\mathrm{temp}}
&=\min_{\mathbf p'\in\mathcal D_f^{\mathrm{pix}}}D_f^{\mathrm{temp}}(\mathbf p'),\\
D_{f,+}^{\mathrm{temp}}&=\max_{\mathbf p'\in\mathcal D_f^{\mathrm{pix}}}D_f^{\mathrm{temp}}(\mathbf p'),\\
\widetilde D_f^{\mathrm{temp}}(\mathbf p)
&=\frac{D_f^{\mathrm{temp}}(\mathbf p)-D_{f,-}^{\mathrm{temp}}}{D_{f,+}^{\mathrm{temp}}-D_{f,-}^{\mathrm{temp}}}.
\end{aligned}
\label{eq:temporal-difference}
\end{equation}
When temporal differences are constant within a frame, we set $\widetilde D_f^{\mathrm{temp}}\equiv1$ to avoid a zero denominator. The event activity map $\widetilde A_f$ uses the same within-frame normalization. We convert the RGB image $I_f=(I_f^R,I_f^G,I_f^B)$ to a grayscale intensity image $Y_f$ and compute normalized edge strength using the standard Sobel kernels $\mathbf B_x^{\mathrm{sob}}$ and $\mathbf B_y^{\mathrm{sob}}$:
\begin{equation}
\begin{aligned}
Y_f&=0.299I_f^R+0.587I_f^G+0.114I_f^B,\\
\nabla_{\mathrm{sob}}Y_f&=(\mathbf B_x^{\mathrm{sob}}*Y_f,\,\mathbf B_y^{\mathrm{sob}}*Y_f),\\
g_f(\mathbf p)&=
\frac{\|\nabla_{\mathrm{sob}}Y_f(\mathbf p)\|_2}
{\max_{\mathbf p'}\|\nabla_{\mathrm{sob}}Y_f(\mathbf p')\|_2+\varepsilon_{\mathrm{num}}}.
\end{aligned}
\label{eq:sobel-edge}
\end{equation}
The grayscale conversion coefficients in Eq. \eqref{eq:sobel-edge} follow ITU-R BT.601~\cite{ref35} and approximate the different human visual responses to red, green, and blue components. Here, $*$ denotes 2D convolution, and $\varepsilon_{\mathrm{num}}$ is a small constant preventing a zero denominator. Larger values of $g_f\in[0,1]$ indicate stronger local appearance boundaries. Normalized temporal differences near the lower and upper endpoints indicate persistent responses and pronounced instantaneous changes, respectively, whereas values near the midpoint are more ambiguous. These quantities enter the temporal-difference reliability weight and candidate score in Eq. \eqref{eq:8}.

\subsection{Depth back-projection and voxel aggregation}
\label{app:projection}

We jointly render depth $D_f(\mathbf p)$ from the decomposed static and dynamic Gaussians. Only candidates with finite depth satisfying $D_f(\mathbf p)>10^{-4}$ are retained. For view $f$ with width $W_f$ and height $H_f$, camera intrinsics are determined from the field of view: $f_x=W_f/[2\tan(\mathrm{FoV}_{x,f}/2)]$ and $f_y=H_f/[2\tan(\mathrm{FoV}_{y,f}/2)]$. The principal point is set to $c_x=W_f/2$ and $c_y=H_f/2$. Thus,
\begin{equation}
\begin{aligned}
\mathbf K_f&=
\begin{bmatrix}
f_x&0&c_x\\
0&f_y&c_y\\
0&0&1
\end{bmatrix},\\[3pt]
\mathbf r_f^{\mathrm{ray}}(\mathbf p)
&=\mathbf K_f^{-1}\widetilde{\mathbf p}
=\begin{bmatrix}
(x-c_x)/f_x\\
(y-c_y)/f_y\\
1
\end{bmatrix},\\[3pt]
\mathbf x_f^{c}(\mathbf p)
&=D_f(\mathbf p)\mathbf r_f^{\mathrm{ray}}(\mathbf p),\\
\mathbf x_f^{w}(\mathbf p)
&=\mathbf R_{cw,f}\mathbf x_f^{c}(\mathbf p)
+\mathbf t_{cw,f}.
\end{aligned}
\label{eq:9}
\end{equation}
Here, $\widetilde{\mathbf p}=(x,y,1)^{\mathsf T}$ denotes homogeneous pixel coordinates, and $\mathbf r_f^{\mathrm{ray}}(\mathbf p)$ is a ray in camera coordinates with its third component fixed at 1. The points $\mathbf x_f^{c}$ and $\mathbf x_f^{w}$ are expressed in camera and world coordinates, respectively. Since rendered depth $D_f(\mathbf p)$ is measured along the camera's optical axis, Eq. \eqref{eq:9} directly scales $\mathbf r_f^{\mathrm{ray}}$ by $D_f$.

The homogeneous extrinsic matrix $\mathbf T_{wc,f}$ maps world coordinates to camera coordinates. For back-projection, we invert it to obtain the camera-to-world transformation. Let $\bar{\mathbf x}=(\mathbf x^{\mathsf T},1)^{\mathsf T}$ denote the homogeneous coordinates of a 3D point. Then,
\begin{equation}
\begin{aligned}
\mathbf T_{cw,f}&=\mathbf T_{wc,f}^{-1},\\
\bar{\mathbf x}_f^{w}(\mathbf p)
&=\mathbf T_{cw,f}\bar{\mathbf x}_f^{c}(\mathbf p),\\
\mathbf R_{cw,f}&=\mathbf R_{wc,f}^{\mathsf T},\\
\mathbf t_{cw,f}&=-\mathbf R_{wc,f}^{\mathsf T}\mathbf t_{wc,f}.
\end{aligned}
\label{eq:camera-to-world}
\end{equation}
Here, $\mathbf R_{wc,f}$ and $\mathbf t_{wc,f}$ denote the rotation and translation of the world-to-camera transformation, while $\mathbf t_{cw,f}$ is the camera center in world coordinates. Transforming event candidates from different views into a common world coordinate system allows their support to be accumulated in the subsequent voxel grid. Depth constraints tie candidates to the current scene geometry, so no new support is generated where valid depth is unavailable.

World-coordinate candidates from Eq. \eqref{eq:9} may overlap across neighboring views and pixels. We group them into regular voxels of side length $h_v$ and use the centroid of the candidates in each voxel as its aggregated position:
\begin{equation}
\begin{aligned}
\mathcal V_v
&=\{\mathbf x_f^w(\mathbf p):\mathbf x_f^w(\mathbf p)\in v\},\\
\mathbf x_v
&=\frac{1}{n_v}
\sum_{\mathbf x\in\mathcal V_v}\mathbf x.
\end{aligned}
\label{eq:voxel-aggregation}
\end{equation}
Here, $\mathcal V_v$ is the set of candidate points in voxel $v$, and $n_v$ is their count. Candidate scores are averaged within each voxel and then normalized across voxels to $[0,1]$, giving the voxel score $\psi_v$. The mean temporal difference determines the dynamic label $\chi_v^{\mathrm{dyn}}$. The final candidate set is selected using Eq. \eqref{eq:10}.

\subsection{Normalization and temporal accumulation}
\label{app:statistics}

The contribution statistics $\kappa_i$, $\bar e_i$, and $m_{2,i}$ are computed by Eq. \eqref{eq:contribution-statistics} in the main text.

The contribution-weighted residual dispersion is $\upsilon_i=\sqrt{\max(m_{2,i}-\bar e_i^2,0)}$. A larger $\upsilon_i$ indicates uneven pixel errors within the projected footprint, with substantially higher residuals in some areas. This usually corresponds to boundary misalignment, local occlusion, or reconstruction anomalies. We therefore define a conservative error $\zeta_i=\operatorname{clip}(\bar e_i+0.25\upsilon_i,0,1)$ that accounts for both mean error and local inconsistency within the footprint.

For the static and dynamic sets separately, we define the scale $\widetilde\kappa$ as the median of current valid contributions, clamped to a minimum of $10^{-6}$. The normalized contribution is $\widehat C_i=1-\exp(-\kappa_i/\widetilde\kappa)$. For Gaussian $i$, let $u$ index valid-observation updates. At its first valid observation, the current statistics initialize $C_i^{(1)}=\widehat C_i^{(1)}$ and $\bar\zeta_i^{(1)}=\zeta_i^{(1)}$. Subsequent updates use an exponential moving average:
\begin{equation}
\begin{aligned}
C_i^{(u)}&=\lambda_{\mathrm{ema}}C_i^{(u-1)}
+(1-\lambda_{\mathrm{ema}})\widehat C_i^{(u)},\\
\bar\zeta_i^{(u)}&=\lambda_{\mathrm{ema}}\bar\zeta_i^{(u-1)}
+(1-\lambda_{\mathrm{ema}})\zeta_i^{(u)}.
\end{aligned}
\label{eq:confidence-ema}
\end{equation}
Here, $\lambda_{\mathrm{ema}}\in[0,1)$ is the smoothing coefficient. The states $C_i^{(u-1)}$ and $\bar\zeta_i^{(u-1)}$ retain the contribution and error from the previous valid update. Both remain unchanged when the current observation is invalid. Omitting the update index and letting $o_i$ denote the number of valid observations, contribution reliability is given by Eq. \eqref{eq:12}.

\subsection{Key selection and canonical neighborhoods}
\label{app:anchors}

A dynamic Gaussian $i$ must satisfy all of the following conditions:
\begin{equation}
C_i\geq\tau_C,\qquad Q_i\geq\tau_Q,\qquad o_i\geq o_{\min},
\label{eq:key-candidates}
\end{equation}
Only then is it included in the candidate set $\mathcal J_{\mathrm{key}}^0$. Here, $\tau_C$ and $\tau_Q$ are contribution and reliability thresholds, respectively, and $o_{\min}$ is the minimum valid-observation count. We define the candidate ranking score as
\begin{equation}
s_i^{\mathrm{key}}=C_i+0.25Q_i+0.02\min(o_i,5),
\label{eq:key-ranking}
\end{equation}
Keys are selected in descending order of $s_i^{\mathrm{key}}$. The constants 0.25, 0.02, and 5 are fixed empirical settings. Contribution is the primary term, while reliability adds at most 0.25 and observation count adds at most 0.10, saturating after 5 observations. This prevents persistently visible Gaussians from dominating the ranking through count alone. These constants are not theoretically derived or probabilistically calibrated. To balance observation quality and spatial coverage, selection first retains the highest-scoring candidate in each voxel. The remaining slots are filled by other high-scoring candidates, preventing Keys from concentrating in a few easily observed regions. The Key count is proportional to the total number of dynamic Gaussians, subject to a minimum target and an upper limit. If $\mathcal J_{\mathrm{key}}^0$ contains too few candidates early in training, selection falls back to all Gaussians using the same score. The resulting Key set is $\mathcal J_{\mathrm{key}}$. Local rigidity constraints preserve the relative geometry of neighboring Keys; these Key--Key links act as motion anchors and provide stable structural references for neighborhood motion. The constraint is implemented by $\mathcal L_{rigid}$ in Eq. \eqref{eq:18}. It preserves lengths of Key nearest-neighbor edges to suppress local structural distortion. Key displacements propagate to Non-key Gaussians along Key--Non-key nearest-neighbor edges, weighted by distance and contribution confidence, forming the CC-guided flow.

For each Non-key Gaussian $i$, we use its canonical-space center to query the $n_{\mathrm{nb}}$ nearest neighbors in the Key set $\mathcal J_{\mathrm{key}}$:
\begin{equation}
\begin{aligned}
\mathcal N_i&=\operatorname{KNN}_{n_{\mathrm{nb}}}
\left(\boldsymbol\mu_i^0;\{\boldsymbol\mu_j^0\}_{j\in\mathcal J_{\mathrm{key}}}\right),\\
d_{ij}&=\left\|\boldsymbol\mu_i^0-\boldsymbol\mu_j^0\right\|_2,
\qquad j\in\mathcal N_i.
\end{aligned}
\label{eq:key-knn}
\end{equation}

\subsection{Auxiliary reconstruction regularizers}
\label{app:regularizers}

The four auxiliary terms in Eq. \eqref{eq:26} are defined as follows:
\begin{equation}
\begin{aligned}
\mathcal L_{dyn}&=\frac{1}{|\mathcal U_f^{\mathrm{dyn}}|}\sum_{\mathbf p\in\mathcal U_f^{\mathrm{dyn}}}
\operatorname{Charb}\bigl(\|\widehat I(\mathbf p)-I(\mathbf p)\|_1\bigr),\\
\mathcal L_{edge}&=\frac{1}{|\mathcal U_f^{\mathrm{edge}}|}\sum_{\mathbf p\in\mathcal U_f^{\mathrm{edge}}}
\|\nabla\widehat I(\mathbf p)-\nabla I(\mathbf p)\|_1,\\
\mathcal L_{art}&=\frac{1}{|\mathcal J_{\mathrm{art}}|}\sum_{i\in\mathcal J_{\mathrm{art}}}\alpha_i,
\quad\mathcal J_{\mathrm{art}}=\{i\mid\bar e_i\ge\tau_e^{\mathrm{art}},\ C_i\le\tau_C^{\mathrm{art}}\},\\
\mathcal L_{mag}&=\frac{1}{|\mathcal J_f^{\mathrm{act}}|}\sum_{i\in\mathcal J_f^{\mathrm{act}}}
\|\Delta\boldsymbol\mu_{i,f}^{\mathrm{hi}}\|_2^2.
\end{aligned}
\label{eq:aux}
\end{equation}
Here, $\operatorname{Charb}(x)=\sqrt{x^2+\varepsilon_{\mathrm{ch}}^2}$ is the Charbonnier function, and $\varepsilon_{\mathrm{ch}}$ is its smoothing constant. The sets $\mathcal U_f^{\mathrm{dyn}}$ and $\mathcal U_f^{\mathrm{edge}}$ denote event-active regions and locally consistent event regions, respectively. The set $\mathcal J_{\mathrm{art}}$ contains Gaussians with persistently high residuals and low contributions. Its residual and contribution thresholds are $\tau_e^{\mathrm{art}}$ and $\tau_C^{\mathrm{art}}$, distinct from the Key selection thresholds. The set $\mathcal J_f^{\mathrm{act}}$ contains currently active Gaussians. Weights $\lambda_d$, $\lambda_b$, and $\lambda_{art}$ are configured per scene. Weights $\lambda_{\mathrm{evt}}(k)$ and $\lambda_{mag}(k)$ vary with training iteration. Appendix \ref{app:parameters} summarizes shared graph and objective weights and identifies scene-dependent weights. The training phases are described in Appendix \ref{app:training}.

\subsection{Baseline availability and comparison scope}
\label{app:baselines}

The remaining event-based methods discussed in the related work were assessed for reproducibility. In our code-availability check as of November 10, 2026, we found no official public code repository or project page for DEGS~\cite{ref41}. E-4DGS~\cite{ref45} had an official repository, but it contained only a README, project images, and similar materials. The authors stated that data and code were still being prepared, and no executable training source code was available. The STD-GS~\cite{ref46} paper page provided the paper and supplementary material, but we found no official code link. We therefore could not include these three methods in evaluation under a common reproduction protocol. E-4DGS also uses multi-view event streams, unlike our monocular setting. Future comparisons will require explicit input adaptation. Event-4DGS is an event-enhanced baseline constructed in the E-D3DGS paper and is distinct from E-4DGS above. The E-D3DGS paper reports better reconstruction quality and rendering speed than Event-4DGS across all twelve scenes~\cite{ref14}. In view of that reported comparison and the lack of an independent Event-4DGS implementation with complete reproduction settings, we exclude it from our unified evaluation and use E-D3DGS as the direct baseline. The previously reported result is not treated as direct comparison evidence under our protocol. These code-availability statements apply only to the date of the check. Methods not included in our evaluation fall outside the scope of our performance conclusions.

\subsection{Training configuration and reporting conventions}
\label{app:training}

\textbf{Implementation details.} Our method and E-D3DGS are implemented in the same codebase and differ only in the modified modules. Both E-D3DGS and ESC-DGR are trained for 50,000 iterations on every scene. In ESC-DGR, static--dynamic decomposition and support augmentation occur at iteration 3,001. Event supervision starts at iteration 8,000. Dynamic densification and pruning end at iteration 18,000, when graph constraints and structural guidance are introduced. High-frequency residuals are gradually incorporated into RGB rendering from iteration 35,000. The threshold network is pretrained for 7,000 steps. Both methods share the warm-up procedure, training inputs, initial Gaussians, decomposition parameters, hardware environment, and random-seed settings. Differences therefore mainly arise from the modules under evaluation. Quality metrics are reported as means and standard deviations over three independent runs with different random seeds; E-D3DGS, ESC-DGR, and all ablation variants use the same three seeds. Cross-scene means are arithmetic averages over the corresponding scenes, and the reported standard deviations are the experimental standard deviations of those aggregate results. The reported domain-level dPSNR gains are calculated from the displayed group means. Recomputing these gains from rounded scene-level values may yield slightly different results. FPS and dynamic Gaussian counts are taken from a single run per scene and aggregated using an arithmetic mean. The prior uses three-frame fusion and a voxel size of 0.01. Rendering contribution statistics are updated every 100 iterations with an EMA coefficient of 0.9. The target Key ratio is 0.04, with a minimum target of 256. We set $\varepsilon_C=0.10$ and $\varepsilon_Q=0.25$. The high-frequency branch has a hidden width of 64 and contains one KAN residual block followed by two MLP residual blocks. The KAN block comprises two layers with zero-order piecewise-constant bases defined over eight segments. Appendix \ref{app:parameters} summarizes the main symbols, shared parameter values, and reported ranges, and identifies scene-dependent settings.

\end{document}